\documentclass{mc}
\renewcommand\datereceived{}
\renewcommand\dateaccepted{}

\usepackage{amsmath, amsthm, amssymb, verbatim, bbm, color, graphics,  geometry}

\usepackage{float}
\usepackage{tabularx}
\usepackage{multirow} 
\usepackage{booktabs}
\usepackage{makecell}
\usepackage{mathrsfs}
\usepackage{caption}
\usepackage{subcaption}
\usepackage{rotating}
\usepackage{braket}
\usepackage{listings}
\usepackage{enumitem}
\usepackage{mathrsfs} 
\usepackage{microtype}
\usepackage{graphicx}
   \usepackage{epsfig}  
\DeclareGraphicsExtensions{.pdf,.png,.jpg,.eps}

\usepackage{booktabs} 
\usepackage{placeins}
\usepackage[normalem]{ulem}

\usepackage{xcolor,sectsty}
\definecolor{astral}{RGB}{46,116,181}
\definecolor{darkslategray}{rgb}{0.18, 0.31, 0.31}
\definecolor{warmblack}{rgb}{0.0, 0.46, 0.36}
\subsectionfont{\color{astral}}
\subsubsectionfont{\color{astral}}
\sectionfont{\color{astral}}
\usepackage{hyperref}
\usepackage{bm}

\newcommand{\mc}{\mathcal}

\begin{document}
	
	\markboth{D. Gupta {\em et al.}}{Wavelet-Accelerated Physics-Informed Quantum Neural Network for Multiscale Partial Differential Equations }
	\title{{\color{warmblack} Wavelet-Accelerated Physics-Informed Quantum Neural Network for Multiscale Partial Differential Equations }}

	\author[D. Gupta {\em et al.}]{{\bf Deepak Gupta}\affil{1}, {\bf Himanshu Pandey}\affil{1} and {\bf Ratikanta Behera}\affil{1}\corrauth}
 
	\address{\affilnum{1}\ {Department of Computational and Data Sciences, Indian Institute of Science, Bangalore, 560012, India}}
	
	\emails{{\tt deepakgupta1@iisc.ac.in} (D. Gupta), {\tt phimanshu@iisc.ac.in} (H. Pandey), {\tt ratikanta@iisc.ac.in} (R. Behera)}
	%

\begin{abstract} 
This work proposes a wavelet-based physics-informed quantum neural network framework to efficiently address multiscale partial differential equations that involve sharp gradients, stiffness, rapid local variations, and highly oscillatory behavior. Traditional physics-informed neural networks (PINNs) have demonstrated substantial potential in solving differential equations, and their quantum counterparts, quantum-PINNs, exhibit enhanced representational capacity with fewer trainable parameters. However, both approaches face notable challenges in accurately solving the multiscale features. Furthermore, their reliance on automatic differentiation for constructing loss functions introduces considerable computational overhead, resulting in longer training times. To overcome these challenges, we developed a wavelet-accelerated physics-informed quantum neural network that eliminates the need for automatic differentiation, significantly reducing computational complexity. The proposed framework incorporates the multiresolution property of wavelets within the quantum neural network architecture, thereby enhancing the network’s ability to effectively capture both local and global features of multiscale problems. Numerical experiments demonstrate that our proposed method achieves superior accuracy while requiring less than five percent of the trainable parameters compared to classical wavelet-based PINNs, resulting in faster convergence. Moreover, it offers three to five times speed-up compared to existing quantum PINNs, highlighting the potential of the proposed approach for solving challenging multiscale and oscillatory problems efficiently.
\end{abstract}

	
\maketitle
\section{Introduction}\label{sec_intro}
Quantum computing has emerged as a powerful computational paradigm with the potential to solve certain problems much faster than classical approaches. A well-known example is Shor's algorithm~\cite{shor1999polynomial}, which enables the factorization of large integers and offers an exponential speedup compared to classical methods. Another key contribution is Grover’s algorithm~\cite{grover1996fast}, which accelerates the search for unstructured data by providing a quadratic improvement over classical algorithms. Further advances, such as quantum phase estimation~\cite{luis1996optimum}, quantum Fourier transform~\cite{coppersmith2002approximate}, and quantum amplitude amplification~\cite{brassard2002quantum}, have laid the foundation for developing quantum algorithms aimed at addressing complex challenges in science and engineering. A significant breakthrough in this direction was the Harrow-Hassidim-Lloyd (HHL) algorithm~\cite{harrow2009quantum}, which provides exponential speedup for solving systems of linear equations under specific conditions. Following the HHL algorithm, several improvements in quantum linear system algorithms (QLSAs) have been developed, including a preconditioned approach for ill-conditioned systems~\cite{clader2013preconditioned}, an algorithm with improved precision scaling~\cite{childs2017quantum}, and the hybrid variational quantum linear solver (VQLS) suitable for near-term devices~\cite{bravo2023variational}. These early implementations provide a strong foundation for tackling problems relevant to real-world applications. A promising domain is scientific computing and machine learning, where the potential advantages of quantum computing are being explored actively~\cite{biamonte2017quantum, Zhao2020VQAs, farhi2018classification}. Many scientific problems involve solving partial differential equations (PDEs), which typically model physical phenomena governed by fundamental laws.

Traditionally, PDEs are solved numerically on classical computers using methods such as the finite difference method (FDM)~\cite{leveque2007} and the finite element method (FEM)~\cite{evans2000}. These methods discretize the computational domain into grid points or elements, reducing the PDE to a system of linear equations, which are then solved numerically at the discretized points. Since classical approaches to linear PDEs ultimately reduce to solving linear systems, it is natural to investigate quantum analogues to exploit potential quantum speedups. Several works in this direction have explored QLSAs for solving linear PDEs~\cite{berry2014high,childs2021high}, with advancements such as preconditioned QLSAs for FEM~\cite{clader2013preconditioned} and quantum fast poisson solvers~\cite{wang2020quantum}. Quantum adaptations of FDM~\cite{cao2013quantum} and VQLS have further enabled applications to both linear and nonlinear systems~\cite{lubasch2020variational}. Quantum algorithms have also been applied to nonlinear PDEs, including Burgers’ and Schrödinger equations~\cite{esmaeilifar2024quantum}.

Most of the above algorithms reduce the problem to a large linear system and utilize QLSAs to obtain the potential speedup. However, these algorithms face significant practical limitations. For example, HHL requires efficient quantum state preparation~\cite{araujo2021divide} and readout, which are challenging. It also demands that the input matrix be sparse and well-conditioned; otherwise, the algorithm's performance deteriorates. Both HHL and VQLS are highly sensitive to noise and hardware errors, and their implementation requires quantum hardware that is far beyond the current capabilities. However, VQLS can solve a large system compared to the HHL algorithm, but it still faces significant challenges when solving nonlinear PDEs\cite{9951265}. Furthermore, converting classical data to quantum states, maintaining precision in quantum phase estimation, and extracting useful information from quantum outputs are other challenges that limit the practical applicability of these quantum solvers. To address these challenges, variational quantum algorithms~\cite{Zhao2020VQAs} and quantum machine learning (QML)~\cite{biamonte2017quantum} have emerged as promising alternatives. These approaches employ parameterized quantum circuits, also known as quantum neural networks (QNNs), which serve as quantum counterparts of classical neural networks. QNNs have been applied to diverse areas such as classification~\cite{farhi2018classification}, supervised learning~\cite{shah2020leveraging}, quantum chemistry~\cite{mcardle2020quantum}, and finance~\cite {pistoia2021quantum}. However, the use of QML in scientific computing, particularly for solving PDEs, is still in its early stages.

Recently, significant progress has been made in the development of classical machine learning approaches for solving PDEs. Among these, the most promising development is physics-informed neural networks (PINNs)~\cite{raissi2019physics}, which offer a mesh-free framework for addressing both the forward and inverse problems governed by PDEs. By embedding the governing physical laws directly into the loss function, which enables neural networks to learn solutions that respect the underlying physics, even when the available data is limited. Despite their advantages, PINNs face challenges when dealing with equations exhibiting sharp gradients or rapid oscillations, and stiff or multiscale PDEs due to varying convergence rates among the loss components~\cite{karniadakis2021physics}. To overcome these challenges, several advancements have been introduced, such as the incorporation of Fourier feature mapping~\cite{tancik2020fourier} to represent high-frequency components in the solution space. Similarly, eigenvector-based representations ~\cite{wang2021eigenvector} have been proposed to capture the multiscale behaviors inherent in many physical systems efficiently. Furthermore, adaptive loss balancing strategies~\cite{mcclenny2023self} have been developed to alleviate issues arising from imbalanced convergence rates among different loss components. Complementary to these efforts, domain decomposition techniques by Wang {\em et al.}~\cite{wang2024practical} have provided a practical means of partitioning complex domains into smaller subdomains, thereby improving the accuracy and computational efficiency in problems with localized features or discontinuities. One recent development is wavelet-based PINNs (W-PINNs)~\cite{pandey2024efficient}, which have been introduced to deal with multiscale PDEs, and they also eliminate the use of auto derivatives from the derivative involved in the loss term. Although classical PINNs have shown significant promise in solving PDEs, their scalability is limited by computational resources and algorithmic challenges. Recently, quantum-based PINNs, by harnessing quantum superposition and entanglement, offer potential speedup and enhanced representational, and the ability to tackle high-dimensional physical systems by leveraging the superposition principle capabilities that are challenging for classical networks, thus opening new frontiers for scientific discovery and engineering applications~\cite{mcardle2020quantum, pistoia2021quantum}.

The rapid development of QML has sparked a wave of research exploring its potential and opened new avenues for scientific computing, particularly in the context of PDEs. Building on the success of PINNs, researchers have begun to explore how QNNs can be integrated into PINN frameworks to leverage quantum computational advantages. The theoretical underpinnings of QNNs as universal function approximators have been investigated \cite{goto2022universal}. Foundational work by Lloyd {\em et al.}~\cite{lloyd2020quantum} provided insights into quantum embeddings for machine learning, which underpins many QNN architectures. QNNs can leverage quantum parallelism and entanglement to achieve potentially superior expressivity and efficiency \cite{du2021learnability, abbas2021power, schuld2021effect}. Recent advancements in physics-informed quantum neural networks (PIQNNs) demonstrate significant progress in addressing PDEs through quantum-enhanced computational frameworks. Early foundational work by Markidis {\em et al.}~\cite{markidis2022physics} established quantum adaptations of PINNs for quantum processing units, demonstrating their applicability to elliptic PDEs such as the Poisson equation through continuous-variable quantum circuits. Subsequent developments bifurcated into purely quantum and hybrid quantum-classical architectures, with Trahan {\em et al.}~\cite{trahan2024quantum} proposing a dynamic quantum circuit integrating Gaussian/non-Gaussian gates within a PINN framework to solve quantum optimal control problems in multi-level systems. A hybrid approach, the hybrid quantum PINN (HQPINN)~\cite{sedykh2024hybrid}, is proposed for nonlinear systems. Architectural innovations have further enhanced parameter efficiency and scalability. Panichi {\em et al.}~\cite{panichi2025quantum} introduced quantum PINNs (QPINNs) that are capable of solving higher-order PDEs. Concurrently, quantum-classical PINNs (QCPINNs) demonstrate parameter reduction compared to classical PINNs \cite{farea2025qcpinn}. The application of PIQNNs has been extended to solve both forward and inverse problems of PDEs~\cite{leong2025hybrid}.

Although the aforementioned developments have demonstrated significant progress, existing approaches still face challenges in accurately capturing sharp gradients, rapid oscillations, and stiff or multiscale PDEs. In addition, most of these methods are highly dependent on automatic differentiation, which imposes a substantial computational overhead. To address these limitations, we propose an accelerated wavelet-accelerated physics-informed quantum neural network (WPIQNN), specifically designed to efficiently address PDEs characterized by sharp gradients, stiff dynamics, highly oscillatory, and multiscale nature of solutions. By representing the solution in terms of wavelet basis functions, our approach mitigates the computational burden by eliminating the need for automatic differentiation required in residual loss and achieves a significant reduction in computational time compared to existing PIQNN frameworks~\cite{leong2025hybrid}. Furthermore, in contrast to W-PINNs, our method significantly reduces the number of trainable parameters and exhibits faster convergence, offering a scalable and efficient framework for solving complex PDE problems. The major contributions of this work are summarized as follows:
\begin{itemize}
    \item We propose a novel WPIQNN approach for solving multiscale PDEs that effectively captures the inherent multiscale nature of solutions.
    \item Our method eliminates the need for automatic differentiation in computing the residual loss, significantly reducing the computational cost compared to existing PIQNN methods.
    \item The proposed method achieves higher accuracy while using less than five percentages of the trainable parameters compared to classical WPINN architectures, leading to faster convergence during training.
    \item Our approach outperforms the existing PINN and PIQNN frameworks in solving multiscale and highly oscillatory PDEs, achieving lower relative $\mathcal{L}^2$-errors and demonstrating more robust convergence.
\end{itemize}
The rest of the paper is organized as follows. Section~\ref{sec_pinn_piqnn} provides background on PINNs and introduces the PIQNN framework, along with its general architecture. Section~\ref{sec_wpiqnn} presents the proposed WPIQNN methodology, including the formulation of the wavelet basis and a detailed description of the designed architecture. Section~\ref{sec_example} reports the numerical experiments to demonstrate the effectiveness of the proposed approach across a variety of problem types. Section~\ref{sec_concl} concludes the paper with a summary of the key findings and potential directions for future research.

\section{Background  and  Related Work}\label{sec_pinn_piqnn}

This section is organized into two subsections. First, we introduce the concept of traditional PINNs along with their extension through wavelet-based architectures, known as Wavelet PINNs. Next, we provide an overview of the PIQNNs approach. We begin by discussing the foundational ideas behind PINNs.

\subsection{Physics-Informed Neural Networks} 
 Physics-informed neural networks, introduced by Raissi~{\em et al.}~\cite{raissi2019physics}, are a class of deep learning models designed to solve PDEs by incorporating the underlying physical laws directly into the training process. Rather than relying solely on data, PINNs embed the governing differential equations, along with the associated initial and boundary conditions, into the loss function of a neural network. The derivatives involved in the loss function were computed efficiently using automatic differentiation.
To illustrate the general PINNs framework, we first consider a general PDE of the form:

\begin{equation}\label{general_pde}
\begin{cases}
\mathcal{N}[u(\bm{x},t)] = f(\bm{x},t), \quad \bm{x} \in \Omega\subset R^{m}, \quad t \in (0, T],\\[6pt]
 \begin{array}{ll}
         u(\bm{x},0)=u_0(\bm{x}), \quad \bm{x} \in \overline{\Omega},\\[6pt]
         \mathcal{B}\left[u(\bm{x},t)\right] = g(\bm{x},t), \quad \bm{x} \in \partial\Omega, \quad t\in (0,T],
     \end{array}
\end{cases}
\end{equation}
where $\mathcal{N}\left[\cdot\right]$ is a nonlinear differential operator, $f$ is a source function, and $\mathcal{B}\left[\cdot\right]$ is a boundary operator, and $u(\bm{x},t)$ is the unknown solution of the equation \eqref{general_pde} defined on $\Omega\times\left[0,T\right].$ To solve this PDE using a PINNs, we approximate $u(\bm{x},t)$ with a neural network $\hat{\mathcal{U}}(\bm{x},t; \theta)$, where $\theta$ are the trainable parameters of the network. For input $\bm{x} \in \mathbb{R}^d$, the network output is computed using the following architecture:
\begin{align*}
    \text{Input layer}: & \quad \mathcal{X}^0(\bm{x},t) = (\bm{x},t), \\
    \text{Hidden layers}: & \quad \mathcal{X}^\ell(\bm{x}) = \sigma(\bm{W}^\ell \mathcal{X}^{\ell-1}(\bm{x}) + \bm{b}^\ell), \quad \ell = 1, \dots, L-1, \\
    \text{Output layer}: & \quad \hat{\mathcal{U}}(\bm{x},t; \theta) = \bm{W}^L \mathcal{X}^{L-1}(\bm{x}) + \bm{b}^L,
\end{align*}
where $\sigma$ is a nonlinear activation function and $\mathcal{X}^\ell$ represents the $\ell^{th}$ layer's output. The training of $\hat{\mathcal{U}}(\bm{x},t; \theta)$ minimizes the composite loss $\mathscr{L}_{\text{total}}$ consists of three components:
\begin{equation}\label{l-total}
     \mathscr{L}_{\text{total}} = \mathscr{L}_{\text{PDE}} + \mathscr{L}_{\text{IC}} +\mathscr{L}_{\text{BC}},
\end{equation}
where
\begin{equation}\label{loss-component}
    \left\{
    \begin{array}{ll}
          \mathscr{L}_{\text{PDE}}=\dfrac{1}{N_c} \displaystyle\sum_{i=1}^{N_c} \left| \mathcal{N}\left[\hat{\mathcal{U}}(x_c^i, t_c^i; \theta)\right] - f(x_c^i) \right|^2,\\[10pt]
          \mathscr{L}_{\text{IC}} = \dfrac{1}{N_0} \displaystyle\sum_{i=1}^{N_0} \left| \hat{\mathcal{U}}(x_0^i, 0; \theta) - u_0(x_0^i) \right|^2,\\[10pt]
          \mathscr{L}_{\text{BC}} = \dfrac{1}{N_b} \displaystyle\sum_{i=1}^{N_b} \left| \hat{\mathcal{U}}(x_b^i, t_b^i; \theta) - g(x_b^i, t_b^i) \right|^2
     \end{array}\right.
\end{equation}

The network parameters $\theta$ are optimized by minimizing $\mathscr{L}_{\text{total}}$ using gradient-based optimizers such as Adam or L-BFGS. The derivatives involved in $\mathcal{N}[\cdot]$ and  $\mathcal{B}[\cdot]$ are computed via automatic differentiation~\cite{baydin2018automatic}. Although this framework provides a flexible and general-purpose approach for solving PDEs, conventional PINNs encounter critical limitations when applied to multiscale problems that exhibit rapid oscillations, sharp gradients, or high-frequency dynamics. These challenges arise primarily from two interconnected factors. First, PINNs are prone to spectral bias, where the neural network tends to favor low-frequency solutions~\cite{deshpande2022investigations}, making it difficult to accurately capture abrupt transitions or high-frequency features. Second, the issue of imbalanced loss terms~\cite{bischof2025multi} often arises where the boundary and initial condition losses may dominate or be dominated by the PDE residual, thereby destabilizing the training process.

To address these limitations, ~\cite{pandey2024efficient} proposed the W-PINN framework, which leverages the multi-resolution capability of wavelet theory\cite{heil1993ten}. In W-PINNs, the solution is approximated in the wavelet space and reconstructed in the physical domain using wavelet basis functions parameterized by dilation and translation. This representation enables localized feature extraction, where finer scales capture detailed structures and translations shift the basis across the domain. A key advantage of W-PINN is that derivatives required in the loss function can be computed analytically using pre-stored wavelet families, avoiding the need for automatic differentiation and reducing computational complexity. Although W-PINN improves training efficiency and accuracy for multiscale PDEs, it still requires a large number of trainable parameters, demanding significant computational resources. Quantum-based approaches offer a promising alternative by reducing parameter requirements while preserving model expressiveness.


\subsection{Physics-Informed Quantum Neural Network}

Recently, hybrid quantum-classical approaches have gained significant attention for solving PDEs by combining the strengths of QNN with classical machine learning techniques. The PIQNN architecture typically consists of three main components. First, an encoding layer is used to encode the input variables into quantum states. Second, a QNN consists of variational layers, entangling gates, and measurement operations to process quantum information and produce classical outputs. Third, a classical optimizer is employed to minimize the loss function by updating parameters $\theta$ of the quantum circuit. These parameters are trained such that the model output aligns with the governing physical laws and supervised data. We start with the first component of the network encoding layer for data encoding.  

\subsubsection{Input data/features encoding}
 In PIQNNs, the first essential step is to encode classical input data, such as spatial and temporal coordinates, into a quantum state. This encoding process plays a crucial role in determining how effectively a quantum model can represent and process problems. There are several types of data encoding available, but two primary quantum data encoding techniques are widely used in QML: angle encoding and amplitude encoding.  The choice of encoding impacts both the expressivity and resource requirements of the quantum model, and is often guided by the dimensionality of the problem and hardware constraints. First, we provide a more detailed explanation of angle encoding.
 \begin{enumerate}[label=(\roman*)]
    \item \textbf{Angle Encoding}: 
    
        Angle encoding is a technique for embedding classical data into quantum circuits by applying parameterized rotation gates to the qubits. To encode a feature vector of length $N$, at least $n$ qubits are required, where $n \geq N$,and each classical feature is mapped to a rotation angle. These rotations can be around the $X$, $Y$, or $Z$ axis, controlled by the choice of gate: $R_X$, $R_Y$, or $R_Z$.
        
        The rotation gates around the $X$, $Y$, and $Z$ axes by angle $\theta$ are given by the following matrices:
        
\begin{equation*}
R_X(\theta) = \begin{pmatrix}
\cos\left({\theta}/{2}\right) & -i \sin\left({\theta}/{2}\right) \\
-i \sin\left({\theta}/{2}\right) & \cos\left({\theta}/{2}\right)
\end{pmatrix},
\quad
R_Y(\theta) = \begin{pmatrix}
\cos\left({\theta}/{2}\right) & -\sin\left({\theta}/{2}\right) \\
\sin\left({\theta}/{2}\right) & \cos\left({\theta}/{2}\right)
\end{pmatrix},
\quad
R_Z(\theta) = \begin{pmatrix}
e^{-i {\theta}/{2}} & 0 \\
0 & e^{i {\theta}/{2}}
\end{pmatrix},
\end{equation*}
respectively. If the number of features is less than the number of qubits, only the first $N$ qubits are rotated, and no operation is applied to the remaining $n-N$ qubits. This encoding method is simple yet powerful and is commonly used in QML and variational algorithms to represent input data within quantum states. This encoding takes advantage of the fact that the probability of measuring a particular state is determined by its phase. Let us consider a single-qubit quantum state $\ket{\psi}$ and a classical parameter $\theta$. A basic angle embedding using $R_Y(\theta)$ rotation on the initial state $\ket{0}$ is given by
\begin{equation*}
    \ket{\psi(\theta, \phi)} = \cos\left(\frac{\theta}{2}\right)\ket{0} + e^{i\phi}\sin\left(\frac{\theta}{2}\right)\ket{1},
\end{equation*}
where the phase factor $e^{i\phi}$ is optional and depends on a specific encoding scheme. More generally, a single-qubit rotation about an arbitrary axis defined by the unit vector $\hat{n} = (n_x, n_y, n_z)$ can be expressed as:
\begin{equation*}
R_{\hat{n}}(\theta) = \exp\left(-i \frac{\theta}{2} (n_x X + n_y Y + n_z Z)\right),
\end{equation*}
where $X$, $Y$, and $Z$ are the Pauli matrices. This formulation allows for flexible encoding of classical data into quantum states via controlled rotations. Although angle encoding is simple and intuitive, it is inefficient for high-dimensional feature spaces because it requires a number of qubits proportional to the number of features. To overcome this limitation, amplitude encoding offers a more compact alternative by encoding data into amplitudes of quantum states. In the following, the concept of amplitude encoding is briefly explained.
\item  \textbf{Amplitude encoding:}
Amplitude encoding is a technique used to encode classical data into the amplitudes of a quantum state. For a normalized classical vector $\mathbf{x} = \left(x_0, x_1, \dots, x_{2^n - 1}\right)^T\in\mathbb{R}^{2^n}$, it can be embedded into the quantum state of $n$ qubits as:
\begin{equation*}
    \ket{\psi(\mathbf{x})} = \sum_{i=0}^{2^n - 1} x_i \ket{i},
\end{equation*}
where $\ket{i}$ are the computational basis states of the $n$-qubit system and the coefficients $x_i$ satisfy the normalization condition $\sum_{i=0}^{2^n - 1} |x_i|^2 = 1$. This form of encoding uses the amplitudes of the quantum state to directly represent classical data, making it exponentially efficient in terms of the number of qubits required. However, preparing such a state generally requires complex quantum circuits or specialized methods such as quantum state preparation algorithms ~\cite{gonzalez2024efficient}. Amplitude encoding is widely used in quantum machine learning and quantum algorithms that represent large vectors in compact quantum form.
\end{enumerate}

\subsubsection{Quantum Neural Network/ Quantum Node}
QNN is the core computational component of the PIQNN framework. It consists of a sequence of parameterized quantum operations, also known as variational quantum layers or trainable layers, followed by entanglement operations and measurements. These components work together to transform encoded quantum states into outputs that can be used for learning tasks. First, we present a brief introduction to variational quantum layers.
 \begin{enumerate}[label=(\roman*)]
     \item \textbf{Variational quantum layers:} 
         These layers are composed of parameterized quantum gates and entangling gates. A parameterized quantum gate is typically a single-qubit rotation such as $R_X(\theta)$, $R_Y(\theta)$, or $R_Z(\theta)$, where $\theta$ denotes a set of trainable parameters. These gates introduce learnable flexibility into the circuit, allowing it to adapt during the training process. Following rotation gates, entanglement is a unique quantum phenomenon that allows qubits to be correlated in ways that are not possible classically. For entangling operations, such as CNOT, CZ, or iSWAP, gates are applied between qubits to introduce quantum entanglement. Unlike classical neural networks, where nonlinearity is introduced via activation functions, quantum circuits leverage the periodic nature of rotation gates and entanglement between qubits to represent complex, high-dimensional functions. Increasing the number of qubits or stacking more variational layers can enhance the model's expressive power, allowing it to approximate a broader class of functions. The specific arrangement and depth of the layers are crucial. Strongly entangled layers, that combine local rotations with dense entanglement, have been found to be effective in a range of PDE problems. Mathematically, we can express this strongly entangled layer as follows for $ n$-qubits:
        \begin{equation*}
            U_l(\theta) = U_{\text{entangle}}\left(\bigotimes_{k=1}^n R_Z(\gamma_{3k})R_Y(\beta_{2k})R_Z(\alpha_{1k})\right),
        \end{equation*}
            where $\bigotimes$ denotes the tensor product and the unitary $U_{\text{entangle}}$ denotes the entanglement operation is composed of a chain of CNOT gates arranged in a circular pattern defined as follows
            \begin{equation*}
            U_{\text{entangle}} = \left(\prod_{k=1}^{n-1} \text{CNOT}_{k,k+1}\right) \text{CNOT}_{n,1}.
            \end{equation*}
\item \textbf{Measurement:}
         After the variational quantum layers have processed the encoded data, the quantum state must be measured to extract classical information that can be used to approximate the solution of the PDE. This is typically achieved by measuring the expectation values of observables, such as the Pauli-Z operator, on the selected qubits. The measurement results yield real-valued outputs, which are then post-processed, often via classical neural network layers, to map them to the desired output dimension.
        \begin{equation*}
            f(\mathbf{x};\boldsymbol{\theta}) = \langle \psi(\mathbf{x})|U^\dagger(\boldsymbol{\theta})MU(\boldsymbol{\theta})|\psi(\mathbf{x})\rangle.
        \end{equation*} 
        Here, $|\psi(\mathbf{x})\rangle$ is the encoded quantum state of input $\mathbf{x}$, and $U(\boldsymbol{\theta})$ is the parameterized variational circuit. The operator $M = \bigotimes_{k=1}^n Z_n$ defines the measurements on the Pauli-Z basis across all qubits. The $f(\mathbf{x};\boldsymbol{\theta})$ yields the expectation value, providing a real-valued output corresponding to the input $\mathbf{x}$.
\end{enumerate}
\begin{figure}[ht]
    \centering
    \includegraphics[width=1\linewidth]{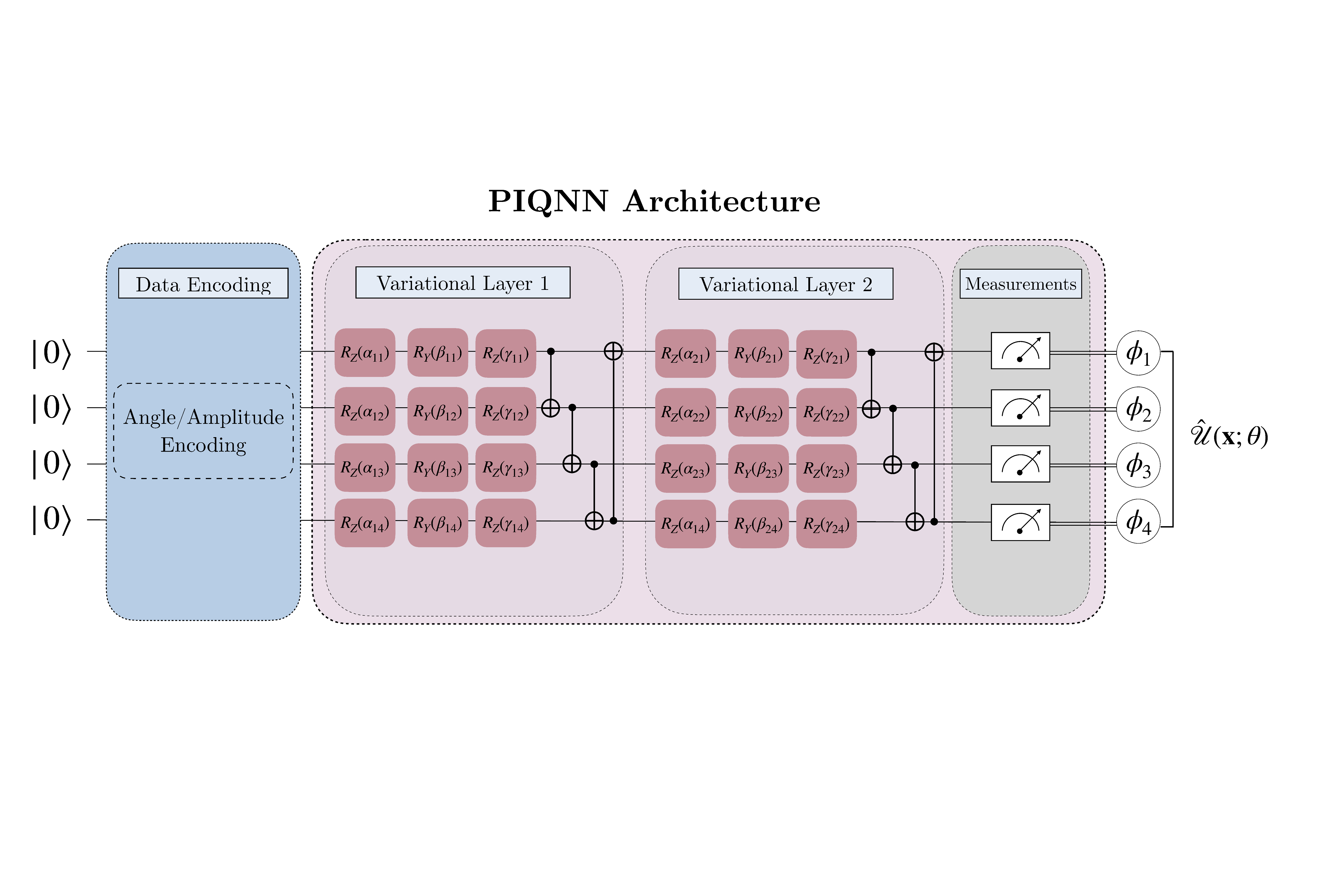}
    \caption{PIQNN Architecture: It consists of an encoding layer followed by variational layers composed of trainable parameters and entangling gates. Final measurements yield the predicted output $\hat{\mathcal{U}}(\boldsymbol{x}; \theta)$. The figure illustrates a 4-qubit setup with two variational layers.
    }
    \label{fig:enter-label}
\end{figure}
\subsubsection{Training and Output Processing of QNNs}
Once the quantum circuit processes the input and produces a measurement outcome, the result is a classical value that must be interpreted and optimized. This is where classical post-processing and optimization loops play a vital role in training the QNN. The measured outputs from the QNN are used to compute a loss function using a physics-informed loss obtained with the help of the governing differential equation and initial and boundary conditions, and minimize the loss. To optimize the trainable parameters $\theta$ of the quantum circuit, a classical Adam or LBFGS optimizer was employed. The classical optimizer iteratively updates the circuit parameters based on the computed loss and its gradient. This hybrid quantum-classical loop continues until convergence. The optimization process uses the PennyLane-based automatic derivatives of hybrid quantum-classical computations \cite{bergholm2018pennylane}.

The PIQNN framework \cite{xiao2024physics} proposes two quantum-enhanced approaches for solving PDEs. First, PIQNN-I processes the inputs through angle encoding, followed by deep variational circuits featuring strongly entangled layers of trainable rotation gates. Measurement via Pauli-Z observables yields classical outputs that approximate the PDE solution after linear transformation. This architecture emphasizes the parameterized quantum circuit depth to enhance the expressivity of general differential equations. Second, PIQNN-II employs an alternating structure that interleaves data re-encoding layers with trainable blocks, thereby enabling a more sophisticated function approximation through repeated feature mapping.

\section{Wavelet-based Physics-Informed Quantum Neural Network (WPIQNN)}\label{sec_wpiqnn}
Traditional PINNs and PIQNNs approximate the solutions of PDEs using activation functions encoded in neural networks. Although these approaches are effective for smooth problems, they often struggle to resolve sharp transitions, localized features, and multiscale phenomena due to their lack of localized representational capacity. Furthermore, PIQNNs/PINNs rely heavily on automatic differentiation using deep networks, which are computationally expensive and numerically unstable in the presence of steep gradients. To address these limitations, we propose a WPIQNN framework that synergizes the multiresolution analysis of wavelets for constructing the solution ansatz,  offering better localization in both space and frequency, and a QNN for better expressibility with the help of fewer parameters compared to its classical version. This approach simultaneously captures both global smooth features through quantum encoding and local sharp variations through wavelet bases, while maintaining the parameter efficiency of quantum representations. Before discussing the detailed architecture, we introduce the basic concept of the wavelets used in this work.

\subsection{Formulation of Wavelets }
Wavelets are localized basis functions that simultaneously capture information in both physical and frequency domains \cite{heil1993ten}. Unlike Fourier or polynomial bases, which are global and often ill-suited for problems with local irregularities, wavelets can represent sharp features with fewer terms due to their inherent compact support. This localization makes them particularly suitable for solving nonlinear PDEs that exhibit phenomena such as shocks or boundary layers \cite{VasilSam14}. In general, a wavelet function is generated from the mother wavelet $\psi(x)$ by scaling and translation as follows:
\begin{equation*}
    \psi_{j,k}(x) = 2^{j/2} \psi(2^j x - k), \qquad \mbox{ for }j,k\in\mathbb{Z},
\end{equation*}
where $j$ denotes the scale and $k$ denotes the translation. Scaling compresses or dilates the wavelet, whereas translation moves it across the domain. For any function $u(x)\in L^2(\mathbb{R})$, we can write the wavelet series expansion as:
\begin{equation*}
    u(x)= \sum_{j=-\infty}^{\infty} \sum_{k =-\infty}^{{\infty }} c_{j,k}\psi_{j,k}(x)
\end{equation*}
where,  $\psi_{j,k}(x)$ are wavelet functions. In order to approximate the solution of a differential equation with the help of a wavelet basis, we have to select the number of family members, for that, we fix the resolution set $\mc{J}= \{J_1, J_2, J_3, \cdots J_N\}\subseteq \mathbb{Z}$. For a given scale \( j = J_i \in \mathcal{J} \), the translation parameter \( k \) ranges from \( \lfloor a \cdot 2^{j+1} \rfloor \) to \( \lceil b \cdot 2^{j+1} \rceil \), where $\lfloor \cdot \rfloor$ and $\lceil \cdot \rceil$  denotes the floor and ceiling function respectively. This method of constructing at wavelet family ensures that the entire spatial domain \([a, b]\) is adequately covered across multiple resolutions. Once the wavelet family is established, we approximate the function $u(x)$ using a finite set of wavelet basis functions as follows:
\begin{equation}
    \hat{u}(x) = \sum_{j = J_1}^{J_N} \sum_{k = \lfloor a \cdot 2^{j+1} \rfloor}^{\lceil b \cdot 2^{j+1} \rceil} c_{j,k} \, \psi_{j,k}(x) + \mathcal{B},
\end{equation}
where
 $J_1$ and $J_N$ denote the minimum and maximum resolution levels in set $\mathcal{J}$, $\psi_{j,k}(x) $ is the wavelet basis function at scale \( j \) and translation \( k \), \( c_{j,k} \) are the corresponding trainable coefficients, and $\mathcal{B}$ is a bias term introduced to improve the expressiveness of the approximation. The extension of wavelet-based approximation to higher dimensions is straightforward. For a two-dimensional function \( u(x, y) \in L^2(\mathbb{R}) \), the wavelet basis functions can be constructed using tensor products of one-dimensional wavelets along the \( x \) and \( y \) directions. A two-dimensional wavelet basis function at scale \( (j_1, j_2) \) and translation \( (k_1, k_2) \) is defined as
\begin{equation*}
    \Psi_{j_1, j_2, k_1, k_2}(x, y) = 2^{j_1/2} \cdot 2^{j_2/2} \, \psi_X(2^{j_1} x - k_1) \, \psi_Y(2^{j_2} y - k_2),
\end{equation*}
where \( \psi_X \) and \( \psi_Y \) are one-dimensional wavelet functions along the \( x \) and \( y \) axes, respectively. To construct a finite wavelet approximation of the function \( u(x, y) \), we choose the resolution sets
$\mathcal{J}_x = \{J_{x1}, J_{x2}, \dots, J_{xN_1}\} \subseteq \mathbb{Z}, \quad \mathcal{J}_y = \{J_{y1}, J_{y2}, \dots, J_{yN_2}\} \subseteq \mathbb{Z}$ for the \( x \) and \( y \) directions. The spatial domain \(\Omega = [a_1, b_1] \times [a_2, b_2] \) is covered by varying translation indices \( k_1 \) and \( k_2 \) according to the resolution. The wavelet approximation of \( u(x, y) \) is then given by
\begin{equation}
    \hat{u}(x, y) = \sum_{j_x = J_{x1}}^{J_{xN_1}} \sum_{j_y = J_{y1}}^{J_{yN_2}} \sum_{k_x = \lfloor a_1 \cdot 2^{j_x+1} \rfloor}^{\lceil b_1 \cdot 2^{j_x+1} \rceil} \sum_{k_y = \lfloor a_2 \cdot 2^{j_y+1} \rfloor}^{\lceil b_2 \cdot 2^{j_y+1} \rceil} c_{j_x, j_y, k_x, k_y} \, \psi_{j_x, j_y, k_x, k_y}(x, y) + \mathcal{B},
\end{equation}
where \( c_{j_1, j_2, k_1, k_2} \) are trainable coefficients and \( \mathcal{B} \) is a bias term.
In our study, we used only the Gaussian wavelet defined as $\psi = - x \, e^{-\frac{x^2}{2}}$. It is a smooth and rapidly decaying structure that provides an efficient representation for functions exhibiting multiscale behavior. It is strongly localized in both the physical and frequency domains, enabling accurate approximation of regions with high gradients or sharp transitions. Additionally, it can capture both low and high-frequency components, ensuring that fine-scale structures are resolved without sacrificing global smoothness. This balance of spatial and spectral resolutions makes the Gaussian wavelet particularly well-suited for representing solutions characterized by scale separation and steep gradients.

\subsection{Methodology}
\begin{figure}[ht]
    \centering
    \includegraphics[width=1\linewidth]{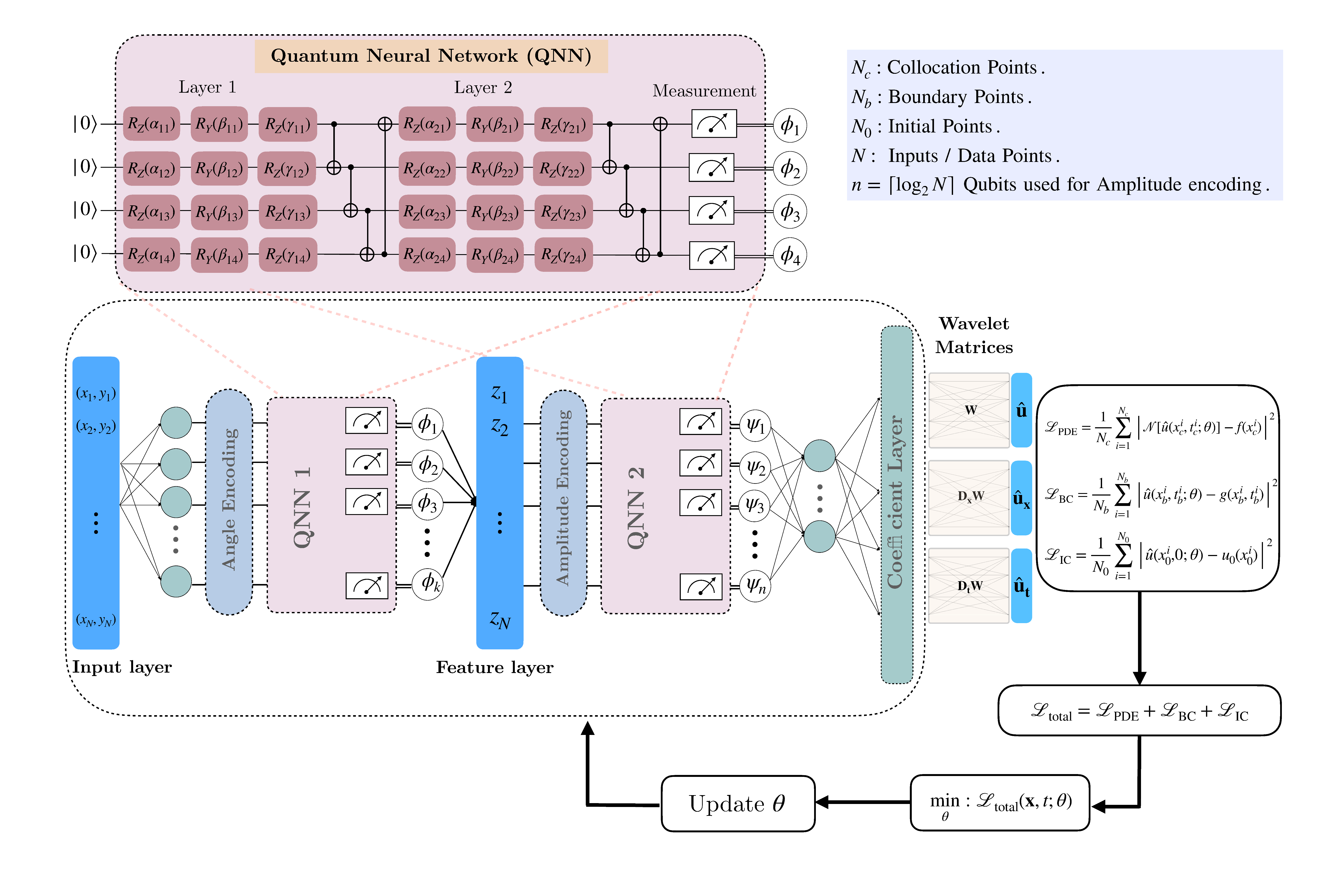}
    \caption{Architecture of the proposed WPIQNN framework. The main components include the input layer comprising spatial-temporal coordinates $(x,t)$, followed by a QNN 1 with angle encoding to extract quantum features. The feature layer is subsequently encoded using amplitude encoding for processing by a QNN 2. The output is passed through classical post-processing layers to produce wavelet coefficients. These coefficients are combined with precomputed wavelet matrices to construct the final loss function.}
    \label{WPQNN_Diagram}
\end{figure}

The architecture of the WPIQNN, depicted in Figure~\ref{WPQNN_Diagram}, was designed to solve the general PDEs of the form presented in Equation~\eqref{general_pde}. The architecture comprises three core components: quantum feature encoder, coefficient prediction, and wavelet-based solution reconstruction.
\FloatBarrier
The process begins by encoding classical input data, specifically the collocation points $(\bm{x}_i, t_i)$, into quantum states using angle encoding. Each input is mapped to a quantum state via parameterized rotation gates, such as $R_Y(x_i)$ and $R_Y(t_i)$, which are distributed over multiple qubits. These quantum states are then passed through QNN 1, which is a trainable parameterized quantum circuit. The output from QNN 1 forms a quantum feature representation $(z_1, z_2, \ldots, z_N)$, which constitutes the feature layer of the WPIQNN architecture. In the case of one-dimensional PDEs, this step can be simplified by directly utilizing the encoded inputs as the feature layer, effectively bypassing QNN 1. Next, the feature layer is encoded into quantum states using amplitude encoding, enabling efficient representation of the feature vector in the amplitudes of the quantum state. These encoded quantum states serve as inputs to a second parameterized quantum neural network, QNN 2, which is also trainable. Upon measurement, the output of QNN 2 yields classical information that passes through the coefficient layer, which maps the quantum outputs to wavelet coefficients $(c_{j,k})$. In the final stage, these coefficients are combined with precomputed wavelet basis functions and their derivative matrices to construct the approximate solution $\hat{u}(\bm{x}, t)$. The derivatives appear in the governing equation to compute the loss of the target PDE. Once we obtain the approximate solution and its derivatives, we can compute the composite loss, which integrates multiple physics-informed components as follows:
\begin{equation*}   
\mathscr{L}_{\text{total}} = \mathscr{L}_{\text{PDE}} + \mathscr{L}_{\text{BC}} + \mathscr{L}_{\text{IC}},
\end{equation*}
This loss is minimized with respect to the trainable parameters $\theta$ of quantum circuits using gradient-based optimization algorithms. The WPIQNN framework requires collocation points spanning the entire spatial and temporal domains for training, along with the wavelet basis matrix and its derivatives up to the order of the differential operator in the PDE. These wavelet matrices serve as fixed, non-trainable components that encode prior knowledge of the solution space, and effectively guide the network to produce physically consistent solutions.

\section{Applications and Discussions}\label{sec_example} 

In this section, we evaluate the performance of the proposed method, WPIQNN, on a series of benchmark problems and compare the results with those obtained using conventional PINNs, WPINN, and other PIQNN approaches. The training process employed the Adam optimizer, which combines momentum-based updates with adaptive learning rates to facilitate convergence in high-dimensional parameter spaces. All neural network weights were initialized using the Xavier initialization to promote a stable gradient propagation during training. For quantitative evaluation, the relative $\mathcal{L}_2$-error is computed over a uniformly sampled validation set, defined as
\begin{equation}
    \mathcal{L}_2 = \frac{\| u_{\theta}(\mathbf{x}) - u_{\text{ref}}(\mathbf{x}) \|_{L^2(\Omega)}}{\| u_{\text{ref}}(\mathbf{x}) \|_{L^2(\Omega)}},
\end{equation}
where $u_{\theta}(\mathbf{x})$ denotes the network prediction and $u_{\text{ref}}(\mathbf{x})$ represents the reference solution. All simulations were conducted using the open-source frameworks \texttt{PennyLane} version $0.40.0$ and \texttt{PyTorch} version $2.6.0$.
 Quantum circuits were constructed and simulated using PennyLane’s built-in \texttt{default.qubit} device, which is a noiseless state-vector simulator designed for ideal quantum circuit simulations on qubit-based architectures. Implementation of the WPIQNN leveraged PennyLane’s quantum node interface. Numerical experiments were performed on an NVIDIA RTX A4500 GPU using CUDA version 12.4. Given the sensitivity of the neural network training to weight initialization, the reported results correspond to the average over 10 independent training runs to ensure statistical robustness. All the hyperparameters used in the experiments are listed in Table~\ref{parameters-table} in Appendix~\ref{appendix}.

\begin{example}[\textbf{High gradient heat conduction problem}]\label{Example-1}
The heat conduction problem in fusion applications involves analyzing rapid temperature variations from sudden heat sources, which are critical for managing the structural integrity of reactors. Its mathematical model, featuring a small parameter $\varepsilon$ that induces sharp thermal gradients, is an effective benchmark for evaluating advanced heat transfer methodologies under extreme conditions. The governing equation for heat conduction is given by
\begin{equation}\label{heat_conduct}
    \left\{
    \begin{array}{ll}
         \dfrac{\partial u}{\partial t} = \varepsilon \dfrac{\partial^2 u}{\partial x^2}+f(x,t), \quad x\in (-1,1),\quad t\in(0,1], \\[8pt]
         u(x,0)=(1-x^2)\exp\left(\dfrac{1}{1+\varepsilon}\right), \quad x\in[-1,1],\\[8pt]
         u(-1,t) = u(1,t)=0, \quad t\in [0,1],
     \end{array}\right.
\end{equation}
where \( u = u(x, t) \) represents the temperature distribution, $t$ is time, $x$ is the spatial domain, $\varepsilon$ is a small positive constant, and $f(x,t)$ is obtained from the exact solution $u(x,t)=(1-x^2)\exp\left(\frac{1}{(2t-1)^2+\varepsilon}\right).$
\end{example}
The behavior of the solution in this heat conduction problem is strongly influenced by the parameter  $\varepsilon$. When $\varepsilon$ is relatively large, solution $u(x,t)$ behaves smoothly throughout the domain.  However, as $\varepsilon$ becomes very small, the solution begins to exhibit sharp gradients and rapid transitions, especially around $t = 0.5$, indicating a clear multi-scale structure. It is important to note that the loss function has two main parts: supervised loss and residual loss. The supervised loss remains small because the boundary values are zero, and the initial values are very small. However, as \( \varepsilon \) decreases, the residual loss, related to the physical equation, starts to increase significantly. This disparity poses a significant challenge for traditional PINNs and existing PIQNNs, which often struggle to capture the dynamics inherent in stiff or multiscale problems. In contrast,  the proposed WPIQNN framework leverages wavelet-based encoding combined with quantum-enhanced layers, which allows for a better resolution of local features and scale-separated structures.

\begin{table}[h]
\centering
\caption{Comparison of different methods for solving the equation\eqref{heat_conduct} with 
 different $\varepsilon$ values}
\begin{tabular}{c|lllc}
\hline\\
$\varepsilon$ & Method &  $\mathcal{L}_2$-error & Trainable parameters\\
\hline\\
[1ex]
\multirow{4}{*}{0.5} 
& CLASSICAL PINN & $1.63\pm 0.75\times 10^{-4}$ &$ 15,501$ \\
& PIQNN-I  \cite{xiao2024physics} & $ 3.41 \pm 1.27 \times 10^{-3} $ &$81 (4 \times 5)$\\
& PIQNN-II  \cite{xiao2024physics} & $4.73 \pm 1.21 \times 10^{-3}$ &$77 (4\times 4)$\\
& W-PINN \cite{pandey2024efficient}&$2.28 \pm 1.21 \times 10^{-5}$ &$15,74,692$\\
& WPIQNN  & $ \bm{1.63 \pm 0.43 \times 10^{-5}}$ &$22,652$\\
[2ex]  
\multirow{4}{*}{0.25} 
& CLASSICAL PINN & $2.27\pm 0.73 \times 10^{-1} $ &$ 15,501$ \\
& PIQNN-I \cite{xiao2024physics} & $ 4.23 \pm 0.37 \times 10^{-2} $ & $81 (4 \times 5)$\\
& PIQNN-II  \cite{xiao2024physics} & $ 3.96\pm 1.12 \times 10^{-2} $ &$97 (5\times 4)$  \\
& W-PINN \cite{pandey2024efficient} &$ 6.26 \pm 1.98 \times 10^{-5} $ &$15,74,692$\\
&  WPIQNN  & $\bm{1.87 \pm 0.37 \times 10^{-5}}$ &$22,652$ \\
[2ex]  
\multirow{4}{*}{0.15} 
& CLASSICAL PINN & $ 1.05 \pm 0.13 \times 10^0 $ & $29,521$ \\
& PIQNN-I  \cite{xiao2024physics} & $ 4.37 \pm 0.87 \times 10^{-1}$ &$97 (4 \times 6)$\\
& PIQNN-II  \cite{xiao2024physics} & $ 2.85 \pm 0.53 \times 10^{-1} $ &$115 (6\times 4)$ \\
& W-PINN     \cite{pandey2024efficient} &$ 3.96 \pm 1.3 \times 10^{-4}$ &$15,74,692$  \\
& WPIQNN    & $\bm{2.13 \pm 0.43 \times 10^{-4}}$ & $22,652$ \\
\hline
\end{tabular}
\label{Heat-table}
\end{table}

A comparative analysis of various methods for solving the heat conduction problem \eqref{heat_conduct} across different values of \( \varepsilon \) is provided in Table~\ref{Heat-table}. The comparison includes metrics such as the relative \( \mathcal{L}_2 \)-error and the number of trainable parameters utilized by each model. For the relatively large value \( \varepsilon = 0.5 \), all methods successfully converge and achieve reasonable accuracy. However, as \( \varepsilon \) decreases to $0.25$ and $0.15$, the performance of classical PINNs and existing PIQNN-(I, II) variants deteriorates significantly. In particular, for \( \varepsilon = 0.15 \), these models fail to accurately approximate the solution, as reflected by large \( \mathcal{L}_2 \)-errors. Although the W-PINN approach demonstrates improved accuracy across all values of \( \varepsilon \), this improvement comes at the cost of a substantially larger number of trainable parameters, approximately 1.5 million. In contrast, the proposed WPIQNN consistently achieves the lowest \( \mathcal{L}_2 \)-errors across all scenarios while requiring only approximately $3$-$5\%$ of the trainable parameters compared to the classical W-PINN. These findings clearly demonstrate that the WPIQNN offers a highly efficient and scalable solution, particularly for stiff or multiscale problems where other models struggle. Its ability to maintain high accuracy with significantly fewer parameters underscores its potential in computationally demanding scientific applications. For all three cases corresponding to \( \varepsilon = 0.5, 0.25, \) and \( 0.15 \), the number of training points was kept consistent across all experiments to ensure a fair comparison. Specifically, number of collocation points was set to \( N_c = 2^{13}\), number of boundary condition points to \( N_b = 1000 \), and number of initial condition points to \( N_{0} = 500 \). Training was performed using the Adam optimizer at a suitable learning rate. In addition, a learning rate scheduler was employed wherever necessary to enhance convergence and training stability. 
\begin{figure}[ht]
    \centering
    \includegraphics[width=0.95\linewidth]{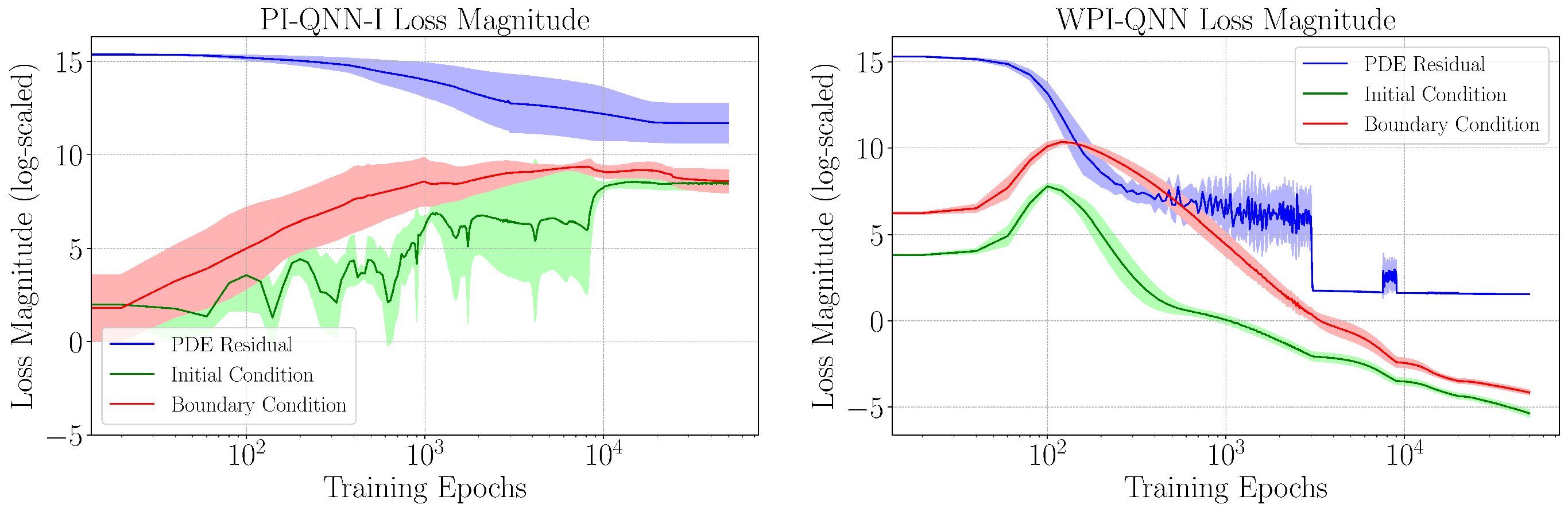}
    \caption{Loss plots with iterations for PIQNN-I (left) and WPIQNN (right) for solving the heat conduction equation~\eqref{heat_conduct} with \( \varepsilon = 0.15 \). Solid lines represent the mean loss, and the shaded areas indicate the corresponding loss variance across 10 independent runs.}
    \label{loss_plot}
\end{figure}
\begin{figure}[ht]
    \centering
    \includegraphics[width=0.95\linewidth]{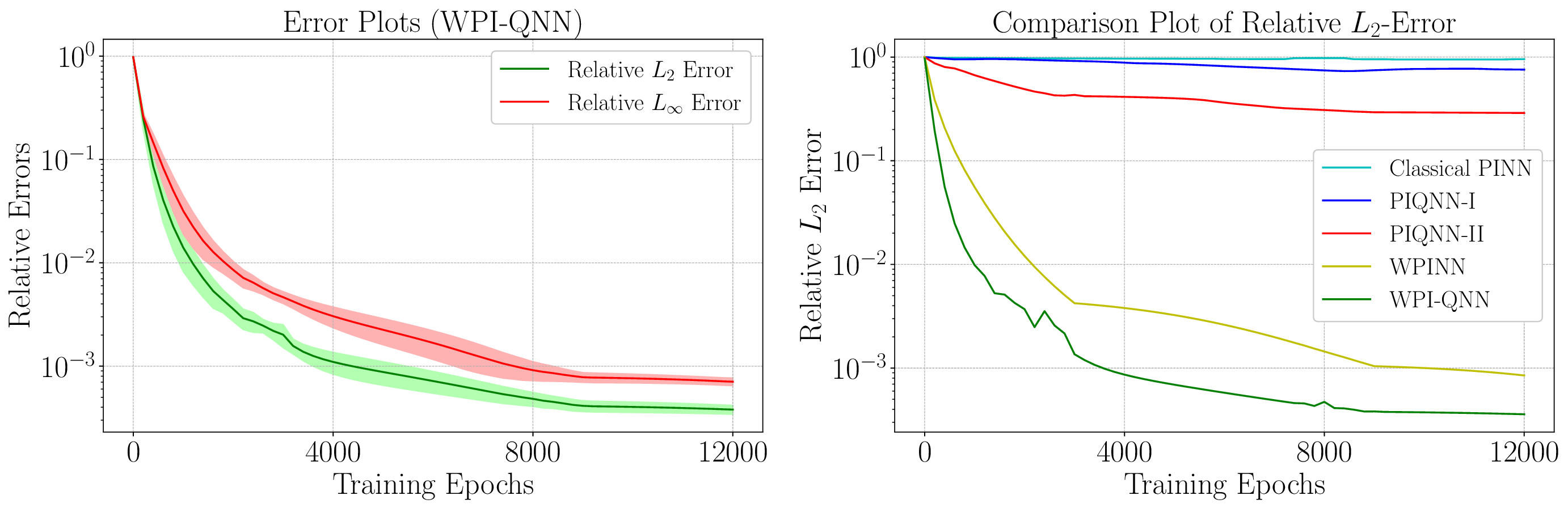}
    \caption{Left: Relative $\mathcal{L}_2$-error with relative \( \mathcal{L}_\infty \)-error WPIQNN. Right: Comparison of relative $\mathcal{L}_2$-error of different methods. Solid lines represent the mean relative $\mathcal{L}_2$-error, and the shaded areas indicate the corresponding error variance across 10 independent runs.}
    \label{HC-error-plot}
\end{figure}
\begin{figure}[ht]
    \centering
    \includegraphics[width=0.95\textwidth]{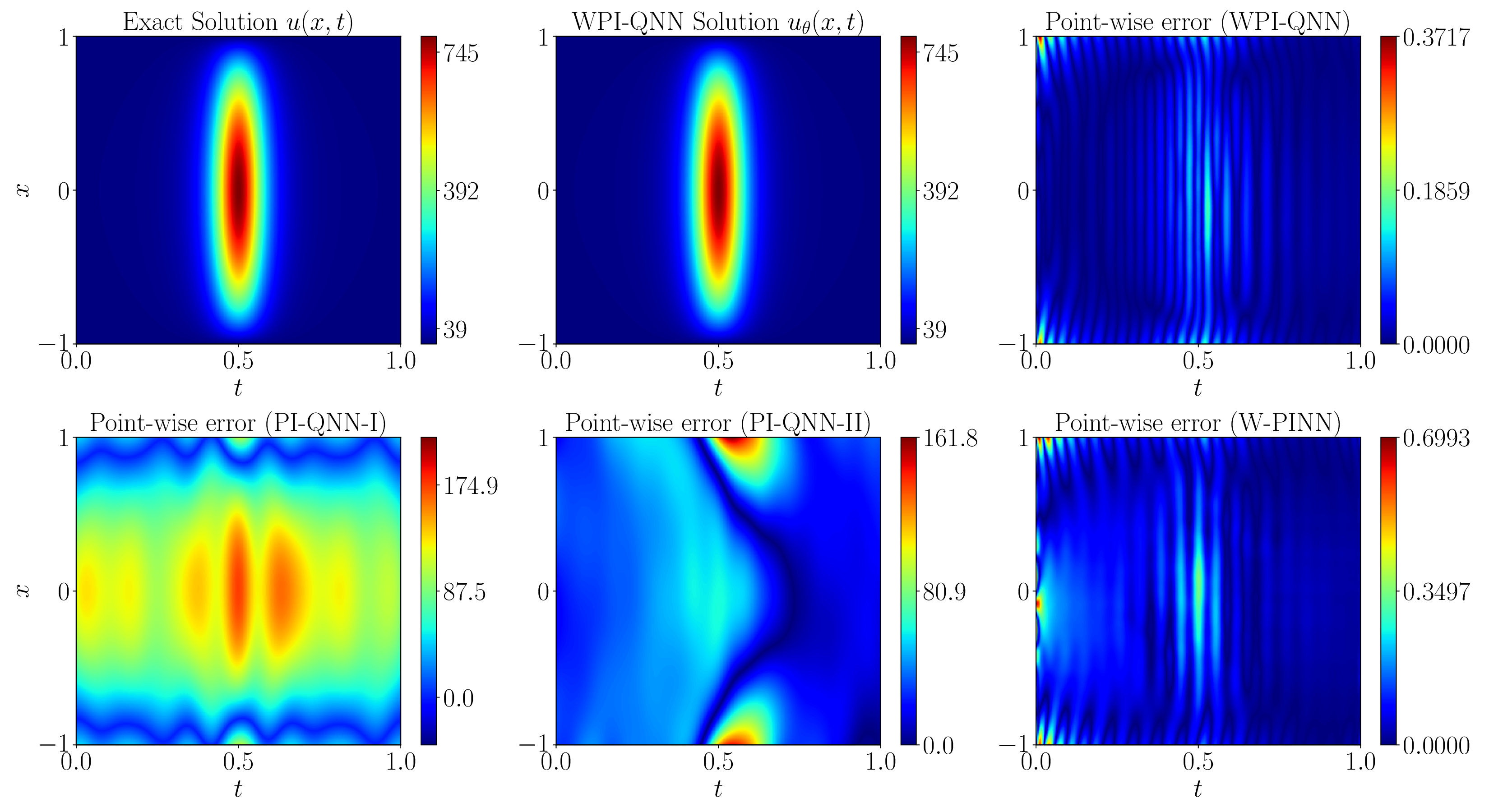}
    \caption{Comparison of exact and predicted solutions for the heat conduction equation~\eqref{heat_conduct} with \( \varepsilon = 0.15 \).\\
    Top row: From left to right-exact solution, WPIQNN prediction, and corresponding point-wise absolute error.\\
    Bottom row: From left to right-pointwise absolute errors for the PIQNN-I, PIQNN-II, and WPINN methods, respectively.}
    \label{HC-imshow-pw}
\end{figure}

The training dynamics of the proposed and baseline models reveal clear distinctions in their ability to enforce underlying physical constraints. As illustrated in Figure~\ref{loss_plot}, PIQNN-I struggles to consistently reduce the PDE residual loss, which remains significantly high even after extended training. Moreover, the losses associated with the initial and boundary conditions (ICs and BCs) exhibit fluctuations and stagnation, indicating poor enforcement of these constraints. In contrast, WPIQNN demonstrates a markedly more stable and efficient training process. All three loss components, PDE, IC, and BC decrease steadily and reach lower magnitudes, with substantially less variance across independent runs. Further evidence of the robustness of the proposed methodology is provided in Figure~\ref{HC-error-plot}, which presents the convergence behavior in terms of the relative $\mathcal{L}_2$-error and relative $\mathcal{L}_\infty$-error. It also presents a comparison of the relative $\mathcal{L}_2$-errors across different methods. It is evident that the WPIQNN achieves superior convergence performance relative to the alternatives, highlighting its enhanced stability and effectiveness in solving the underlying problem.

As shown in Figure~\ref{HC-imshow-pw}, the proposed WPIQNN framework consistently achieves significantly lower pointwise absolute errors across the entire domain compared to other methods. In contrast, both PIQNN-I and PIQNN-II display substantial inaccuracies, with visibly larger error magnitudes distributed throughout the domain. These elevated errors suggest difficulties related to convergence and suboptimal training dynamics inherent in the existing PIQNN architectures. Although WPINN demonstrates relatively improved performance over the PIQNN variants, it still fails to accurately capture the solution near the left boundary and around $t = 0.5$, where the reference solution exhibits a sharp gradient. In contrast, the WPIQNN not only yields smaller absolute errors overall but also maintains robustness in accurately approximating regions of rapid variation.

\begin{table}[ht]
    \renewcommand{\arraystretch}{1.4}
    \centering
    \caption{$\mathcal{L}_2$-errors of the WPIQNN method for Equation\eqref{heat_conduct} with different $\varepsilon$.}
    \begin{tabular}{ccccc}
        \hline
        $\varepsilon$ & $\mathcal{L}_2$-error & $N_c$ & $ N_b$ & $N_0$ \\
        \hline
        0.14 & $5.34\pm 1.32 \times 10^{-4}$ & $2^{14} $& $ 2000$ &   $1000$\\
        0.13 & $2.27\pm 0.55 \times10^{-4} $ & $2^{14} $& $ 2000$ &   $1000$\\
        0.12 & $6.54\pm 0.95 \times10^{-4} $ & $2^{14} $& $ 2000$ &   $1000$\\
        0.11 & $1.17\pm 0.57 \times10^{-4}$ & $2^{15} $& $ 8000$ &   $4000$\\
        \hline
    \end{tabular}
    \label{HC-small}
\end{table}
We further assess the performance of the proposed WPIQNN method for smaller values of \( \varepsilon \), where the multiscale nature of the heat conduction problem becomes increasingly pronounced. The results, summarized in Table~\ref{HC-small}, show the relative \( \mathcal{L}_2 \)-errors achieved by WPIQNN for \( \varepsilon = 0.11, 0.12, 0.13, \) and \( 0.14 \). As \( \varepsilon \) decreases, the problem exhibits sharper gradients and highly localized variations, substantially increasing the difficulty level. In particular, for \( \varepsilon = 0.11 \), the solution undergoes drastic changes near \( t = 0.5 \), varying from $0$ to approximately $9000$, which clearly reflects the highly multiscale nature of the problem. Despite this increased complexity, the WPIQNN model consistently delivers accurate and stable predictions, achieving relative \( \mathcal{L}_2 \)-errors of the order of \( 10^{-4} \). In these challenging regimes, the other aforementioned methods failed to converge or provide meaningful solutions. Therefore, results are reported solely for our proposed WPIQNN, as it is capable of accurately capturing the intricate solution structures while maintaining stability across highly stiff and multiscale scenarios. These findings further highlight the robustness and effectiveness of WPIQNN in addressing problems where existing approaches struggle.

\begin{figure}[ht]
    \centering
    \includegraphics[width=0.95\linewidth]{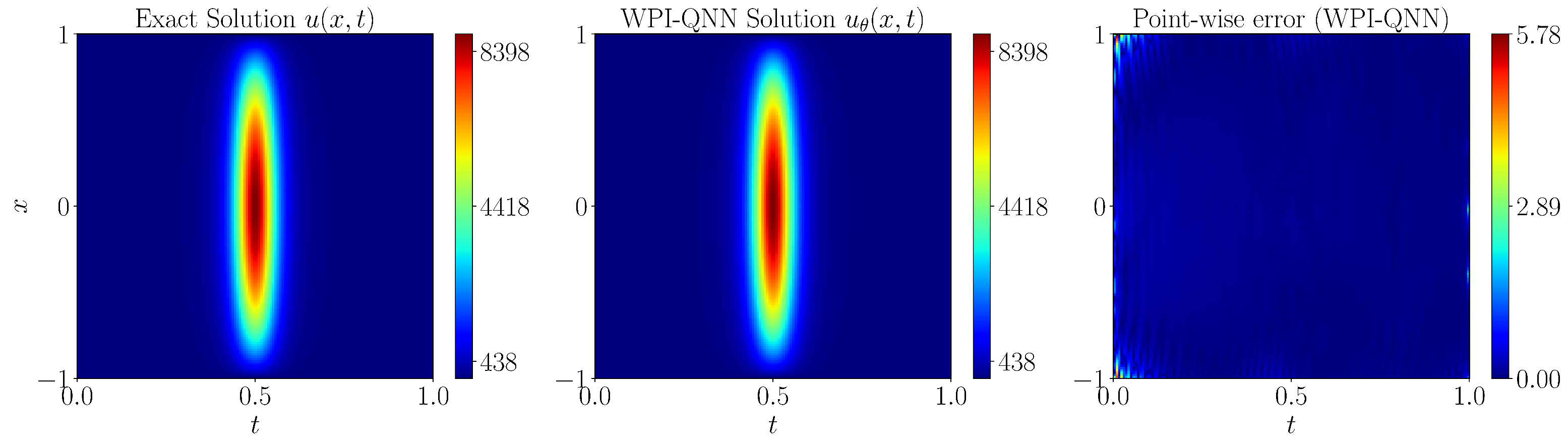}
    \caption{Exact solution $u(x,t)$ (left), WPIQNN-predicted solution $u_\theta(x,t)$ (middle), and the corresponding point-wise error (right) for the equation~\eqref{heat_conduct} with $\varepsilon=0.11$.}
    \label{HC-Eps11}
\end{figure}
The effectiveness of the proposed WPIQNN method is further demonstrated in Figure~\ref{HC-Eps11}, which shows the numerical solution for the heat conduction problem~\eqref{heat_conduct} with \( \varepsilon = 0.11 \). In this case, the solution exhibits extreme variations, particularly near \( t = 0.5 \), where the values transition sharply from near zero to approximately $9000$ within a narrow temporal region. Such rapid changes present significant challenges for conventional neural network-based solvers, which often struggle to capture localized and steep gradients accurately. Indeed, existing methods, including PINNs, PIQNN-I, PIQNN-II, and WPINN, consistently failed to produce stable or accurate approximations for this case, frequently exhibiting either divergence or stagnation during training. In contrast, the proposed WPIQNN approach successfully captures the intricate solution structure, maintaining uniformly low point-wise absolute errors across the entire spatial and temporal domain. Quantitatively, WPIQNN achieves a relative \( \mathcal{L}_2 \)-error of \( 1.17 \pm 0.57 \times 10^{-4} \), highlighting its robustness and superior accuracy in addressing highly stiff and multiscale problems.

\begin{figure}[ht]
    \centering
    \includegraphics[width=0.85\linewidth]{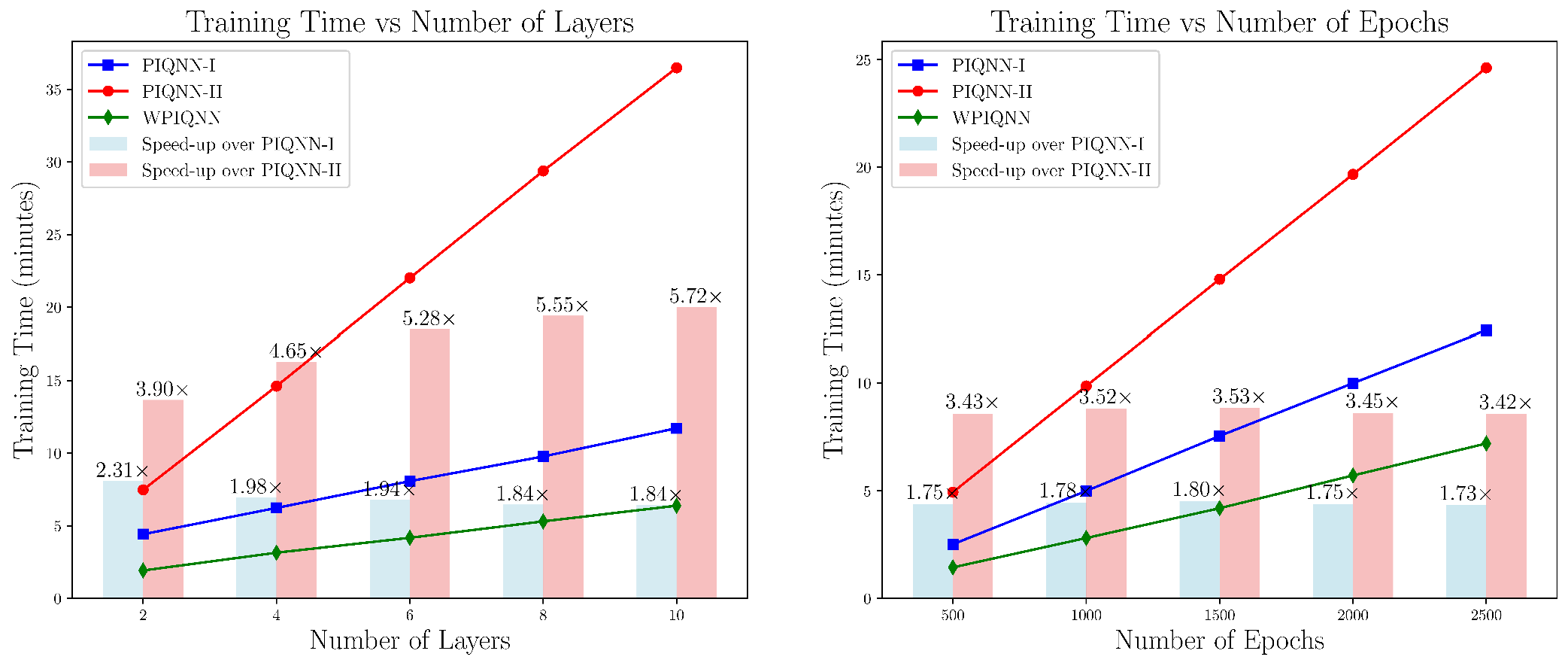}
    \caption{Time comparison of WPIQNN with PIQNN-I and PIQNN-II. Layers vs Time (Left), Epochs vs Time (Right). Solid lines indicate training time and bars represent speedup achieved by WPIQNN.}
    \label{time-comp}
\end{figure}
To further demonstrate the computational efficiency of the proposed WPIQNN architecture over existing quantum-inspired approaches PIQNN-(I, II), we conduct a detailed comparison of training times for solving the one-dimensional heat conduction problem~\eqref{heat_conduct} with \( \varepsilon = 0.25 \). The results are presented in Figure~\ref{time-comp}, where two key scenarios are considered, first varying the number of layers (left) and second varying the number of training epochs (right). For the first case, training was performed using the Adam optimizer over a fixed $1000$ epochs and other training configurations follow those used in the previous examples to ensure consistency. The plots use green, blue, and red lines to represent WPIQNN, PIQNN-I, and PIQNN-II, respectively, while light blue and light red bars denote the ratios of PIQNN-I/WPIQNN and PIQNN-II/WPIQNN training times. As illustrated, the proposed method consistently achieves shorter training times as the number of layers increases, yielding approximately a $2$ times speed-up over PIQNN-I and up to a $5$ times speed-up compared to PIQNN-II. In the second case, with the number of layers fixed at $4$ and varying the number of epochs, a similar trend is observed: the WPIQNN architecture maintains a roughly $2$ times reduction in training time over PIQNN-I and around $3.5$ times over PIQNN-II across all configurations. These improvements are primarily attributed to the removal of automatic differentiation in computing the residual loss, which significantly reduces computational overhead.

Although no direct comparison is provided with the classical W-PINN architecture, quantum circuit simulations on classical hardware remain computationally expensive despite having fewer trainable parameters. This is because of the exponential scaling of quantum states and the complexity of matrix operations, whereas classical neural networks benefit from highly optimized GPU-based frameworks. A fair assessment of computational advantages can only be made once the proposed method is deployed on actual quantum hardware, where quantum parallelism is expected to further enhance the performance. The current results clearly highlight the superior time efficiency of WPIQNN over other quantum-based approaches.

\begin{example}[\textbf{Helmholtz equation in high-frequency regime}]\label{Example-2}
    The Helmholtz equation is a fundamental elliptic PDE that arises in the modeling of wave propagation phenomena, including acoustics, electromagnetics, and optics. It typically takes the following form:
\begin{equation}\label{Helm-holts}
\begin{cases}
\Delta u(x, y) + \kappa^2 u(x, y) = f(x, y), & (x, y) \in \Omega = (-1, 1)^2, \\[8pt]
u(x, y) = g(x, y), & (x, y) \in \partial\Omega,
\end{cases}
\end{equation}
where $\Delta$ denotes the Laplacian operator, $\kappa$ represents the wave number (related to frequency by $\kappa = \omega/c$), and $f$ and $g$ are the source and boundary functions, respectively.
\end{example} In high-frequency domains, the Helmholtz equation becomes increasingly difficult to solve using traditional numerical methods due to the rapid oscillations in its solutions. To benchmark our model, we consider an exact solution of the form: $ u(x, y) = \sin(\pi b_1 x) \sin(\pi b_2 y),$ from which functions $f(x, y)$ and $g(x, y)$ are derived accordingly. In our simulations, we set $\kappa = 1$, $b_1 = 1$, and $b_2 = 8$. To ensure a fair evaluation, both models were trained using the Adam optimizer with a learning rate scheduler. The number of collocation points was fixed at \(N_c = 2^{14}\), and boundary condition points at \(N_b = 1000\). Each architecture employed two hidden layers in the feature mapping network and four layers in the secondary neural network. A wavelet basis resolution of \([-4, 5]\) was used consistently across both models.
\begin{table}[H]
\centering
\caption{Comparison of different methods for solving Helmholtz equation\eqref{Helm-holts}}
\label{tab:performance}
\begin{tabular}{lccccc}
\toprule
Method &  $\mathcal{L}_2$-error  & Trainable parameters \\
\midrule
Classical-PINN  & $2.39 \pm 2.17 \times 10^{-1}$  & $15,501$ &  \\
PIQNN-I \cite{pandey2024efficient} & $9.12\pm 1.71 \times 10^{-1}$  & $197 (8\times7)$ \\
PIQNN-II \cite{pandey2024efficient} & $3.12\pm 0.71 \times 10^{-1}$  & $126 (6\times5)$ \\
WPINN \cite{pandey2024efficient} & $3.12\pm 0.71 \times 10^{-4}$  & $22,27,696$ \\
WPIQNN &  $\mathbf{2.19\pm 0.43 \times 10^{-4}}$  & $35,611$\\
\bottomrule
\label{HelmHoltz-table}
\end{tabular}
\end{table}
A comparative analysis of the proposed WPIQNN method for solving the Helmholtz equation~\eqref{Helm-holts} with PIQNN-I, PIQNN-II, classical PINN, and WPINN is presented in Table~\ref{HelmHoltz-table}. The comparison focuses on two primary metrics: the relative \(\mathcal{L}_2\)-error and the number of trainable parameters. The results show that WPIQNN consistently outperforms the other methods, achieving comparable or superior relative \(\mathcal{L}_2\)-errors. It is also observed that the WPIQNN requires only $35,611$ parameters, which is approximately $5\%$ of the parameters used by the WPINN ($2,227,696$ parameters). Although PINN, PIQNN-I, and PIQNN-II use fewer parameters, they consistently fail to converge under the same training configuration. These results highlight the exceptional parameter efficiency of WPIQNN, which achieves high accuracy with substantially fewer parameters.

\begin{figure}[ht]
    \centering
    \includegraphics[width=0.85\linewidth]{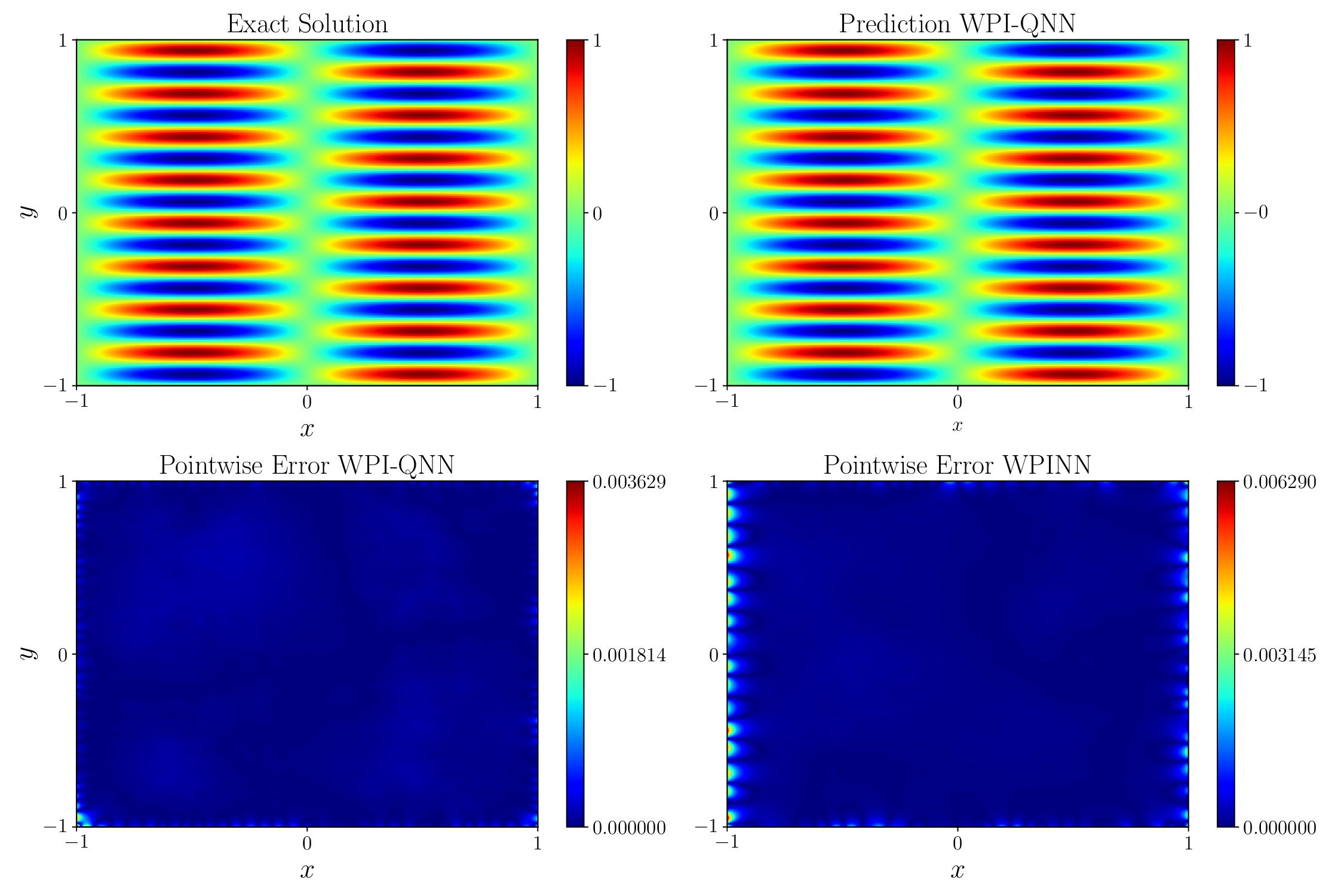}
    \caption{Solution and error visualization for the 2D high-frequency Helmholtz equation \eqref{Helm-holts}. Top: Exact solution (left) and WPIQNN prediction (right), showing accurate reconstruction of high-frequency features. Bottom: Pointwise absolute errors for WPIQNN (left) and WPINN (right), with WPIQNN exhibiting lower error across the domain, especially near boundaries.}
    \label{helmholtz-pw}
\end{figure}
The superior predictive capability of the proposed WPIQNN framework for solving the Helmholtz equation~\eqref{Helm-holts} is evident from the results presented in Figure~\ref{helmholtz-pw}. The visualization highlights the ability of WPIQNN to accurately reconstruct the underlying oscillatory structures of the solution, effectively capturing the high-frequency components that often present significant challenges for conventional PINN-based approaches. Moreover, WPIQNN achieves a substantial reduction in point-wise absolute errors compared to WPINN, consistently maintaining low error magnitudes across the entire domain, including near-boundary regions where classical models typically suffer from accuracy deterioration. These findings demonstrate the enhanced accuracy and computational efficiency of WPIQNN in solving complex PDEs, particularly in high-frequency regimes, thereby reinforcing its advantages over traditional neural network-based methods.

\begin{figure}[ht]
    \centering
    \includegraphics[width=0.85\linewidth]{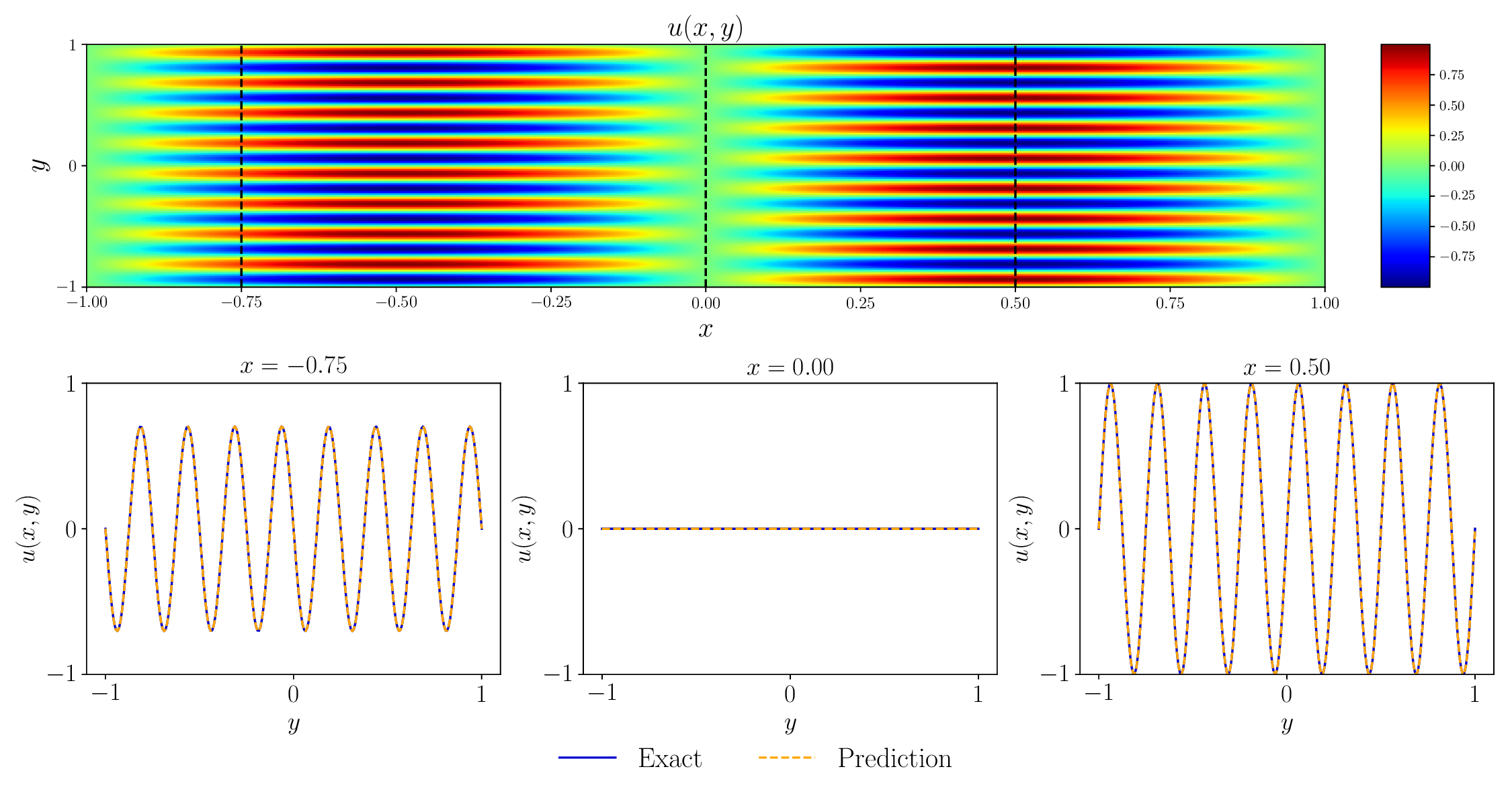}
    \caption{ Comparison of the cross-sections of the predicted and exact solutions for various $x-$domain snapshots for the Helmholtz equation\eqref{Helm-holts} using WPIQNN method at location ($-0.75$, $0.0$, and $0.5$).}
    \label{Helmholtz-snap}
\end{figure}
Effectiveness of WPIQNN in accurately approximating the sinusoidal solution of the 2D Helmholtz equation~\eqref{Helm-holts} is illustrated through the comparison of individual cross-sections of the predicted and exact solutions, as shown in Figure~\ref{Helmholtz-snap}. The top panel presents the predicted solution $u(x,y)$ obtained via WPIQNN, whereas the bottom panels provide 1D cross-sectional plots at $x = -0.75$, $0.0$, and $0.5$. These cross-sections clearly demonstrate excellent agreement between the predicted and exact solutions across the domain, highlighting the accuracy and robustness of the proposed method.


\begin{example}[\textbf{Klein-Gordan equation}]\label{Example-3}
The Klein-Gordon equation, which originates from relativistic quantum mechanics, is a second-order partial differential equation that generalizes the classical wave equation. Due to its significance in physics, it has been extensively explored from both analytical and numerical perspectives. The one-dimensional form of the Klein-Gordon equation is:
\begin{equation}\label{KG-eqn}
\begin{cases}
u_{tt} + \alpha u_{xx} + \beta u + \gamma u^k = f(x,t), \quad (x,t) \in \Omega \times (0,T], \\[6pt]
u(x,0) = x, \quad x \in \Omega,  \\[6pt]
u_t(x,0) = 0, \quad x \in \Omega, \\[6pt]
u(x,t) = h(x,t), \quad (x,t) \in \partial \Omega \times [0,T].
\end{cases}
\end{equation}
\end{example}
To solve this problem, we choose $\Omega = [0,1]$ and $T=1$. The parameters are set as $\alpha = -1$, $\beta = 0$, $\gamma = 1$, and $k = 3$. The exact solution for this problem is given as follows
\begin{equation}\label{KG_exact_sol}
    u(x,t) = x \cos(a \pi t) + (x t)^3,
\end{equation}
As the parameter \(a\) in the exact solution ~\eqref{KG_exact_sol} increases, the classical PINNs and PIQNNs approaches struggle to produce reliable approximations due to the growing complexity of the underlying oscillations. This difficulty is often attributed to imbalanced gradient contributions from different terms in the loss function during training, a phenomenon known as gradient imbalance. Consequently, classical PINN fail to converge or yield large approximation errors in such regimes. These challenges make the Klein–Gordon equation an appropriate testbed for evaluating the effectiveness of the proposed WPIQNN architecture.

For the training setup in both cases \(a = 5\) and \(a = 10\), the WPIQNN and WPINN architectures utilized the same network configuration, comprising two hidden layers in the feature mapping network and four layers in the secondary neural network. However, the collocation points and wavelet resolution are adapted to account for the increasing complexity of the solution. Specifically, for \(a = 5\), the number of collocation points was set to \(N_c = 2^{13}\), with \(N_0 = 500\) initial points and \(N_b = 1000\) boundary points, and a wavelet resolution of \([-10,4]\) was employed. The model was trained using the Adam optimizer for \(100\)k epochs, starting with a learning rate of \(10^{-2}\), which was reduced by an order of magnitude at 10k, 40k, and 80k epochs, with a minimum of \(10^{-5}\). For the more challenging case of \(a = 10\), the number of collocation points was increased to \(N_c = 2^{14}\) to adequately resolve the highly oscillatory solution, and the wavelet resolution was extended to \([-10,5]\). Training was conducted for \(400\)k epochs with a similar learning rate decay schedule at 10k, 100k, and 200k epochs.
\begin{table}[ht]
\centering
\caption{The performance of different methods for solving equation~(\eqref{KG-eqn}).}
\begin{tabular}{lcccc}
\toprule
\multirow{2}{*}{\textbf{Method}} & \multicolumn{2}{c}{$\mathbf{a=5}$} & \multicolumn{2}{c}{$\mathbf{a=10}$} \\
\cmidrule(lr){2-3} \cmidrule(lr){4-5}
 & $\mathcal{L}_2$-error &  Trainable parameters  & $\mathcal{L}_2$-error & Trainable parameters  \\
\midrule
Classical PINN & $5.73 \pm 2.79 \times 10^{-1}$ & $61,001$ & $9.17 \pm 2.57 \times 10^{-1}$ &$ 61,001$ \\
PIQNN-I \cite{xiao2024physics} & $9.13\pm 1.27 \times 10^{-1}$  & $197 (8\times7)$ & $1.18\pm 2.17 \times 10^{0}$  & $197 (8\times7)$ \\
PIQNN-II \cite{xiao2024physics} & $7.13\pm 1.27 \times 10^{-1}$  & $126 (6\times5)$ & $1.18\pm 2.17 \times 10^{0}$  & $126 (6\times5)$ \\
W-PINN \cite{pandey2024efficient} & $2.53\pm 1.87 \times 10^{-3}$  & $9,14,956$ & $4.18\pm 2.17 \times 10^{-3}$  & $34,16,627$ \\
WPIQNN &  $\mathbf{9.65\pm 2.17 \times 10^{-4}}$  & $16,596$ & $\mathbf{1.72\pm 1.57 \times 10^{-3}}$  & $50,837$ \\
\bottomrule
\end{tabular}
\label{KG-table}
\end{table}


\begin{figure}[ht]
    \centering
    \includegraphics[width=0.95\linewidth]{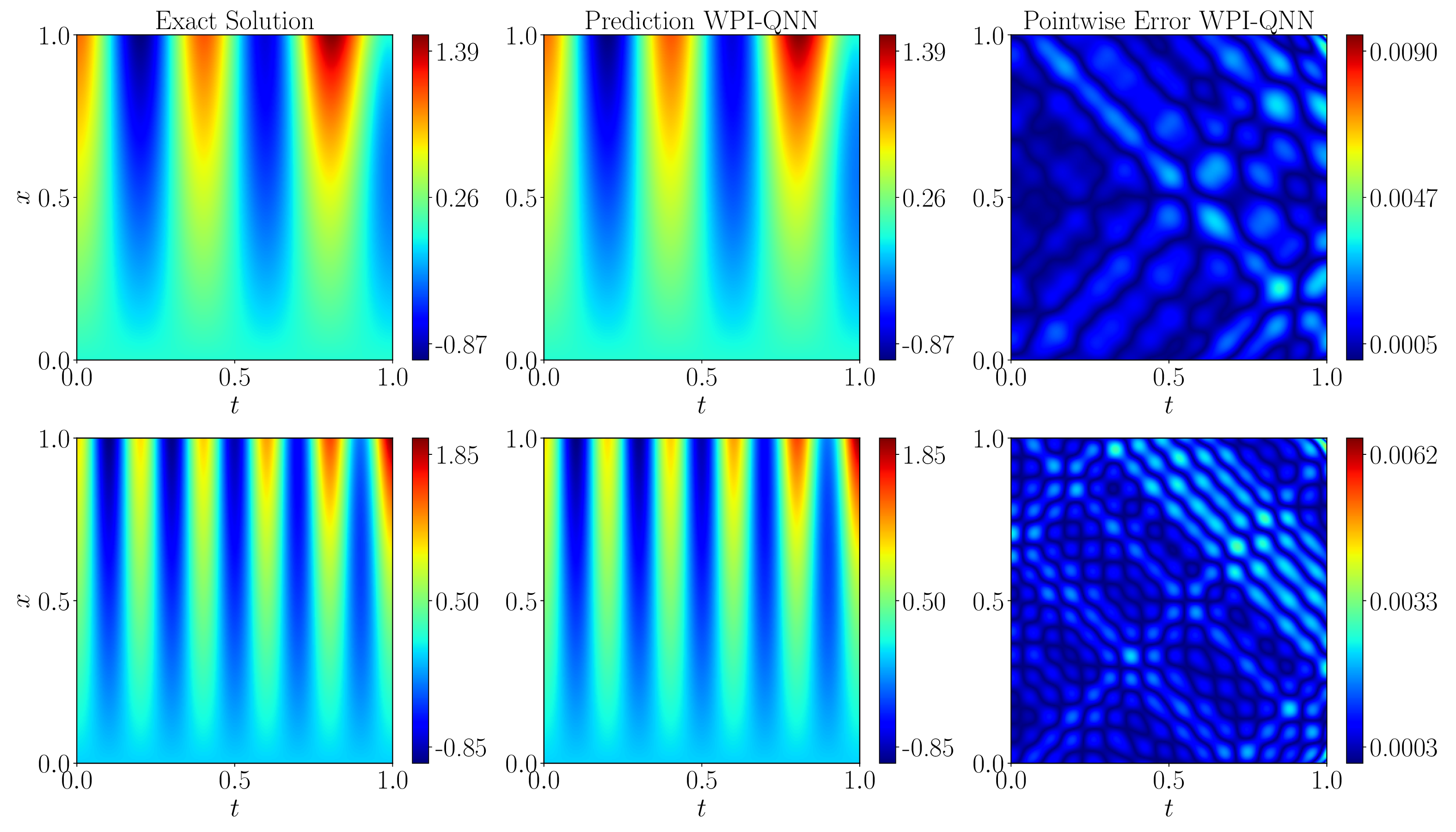}
    \caption{
        WPIQNN solution of the 1D Klein–Gordon equation~\eqref{KG-eqn} for $a=5$ (top row) and $a=10$ (bottom row). 
        From left to right in each row: Exact solution, WPIQNN prediction, and corresponding pointwise absolute error.}
    \label{KG-Exact-pred}
\end{figure} 

Table~\ref{KG-table} presents a comparison of the proposed WPIQNN framework with the classical PINN and WPINN methods for solving the 1D Klein-Gordon equation~\eqref{KG-eqn}. The results are reported for two cases, corresponding to different values of the parameter \(a\) in the exact solution~\eqref{KG_exact_sol}, specifically for \(a=5\) and \(a=10\), to evaluate the models under varying levels of difficulty. As summarized in Table~\ref{KG-table}, the proposed WPIQNN consistently outperforms PIQNN-(I, II), classical PINN, and WPINN in terms of accuracy while achieving a remarkable reduction in the number of trainable parameters, using only $2$--$5\%$ of the parameters required by WPINN. For \(a=5\), the proposed WPIQNN achieves a significantly lower relative \(\mathcal{L}_2\) error compared to PIQNN-(I, II), and classical PINN by a large margin, and also provides a better relative \(\mathcal{L}_2\) error than WPINN, while using only a few trainable parameters, whereas the PIQNN-(I, II), and classical PINN struggle to provide an accurate approximation, and provide a large relative error of order \(10^{-1}\). For the more challenging case of \(a=10\), in which the solution contains higher-frequency oscillations, the performance gap between the methods becomes even more apparent. The WPINN, despite employing over three million parameters, achieves only comparable relative \(\mathcal{L}_2\)-errors. In contrast, WPIQNN not only achieves superior accuracy but also uses only approximately $2\%$ of the parameters required by classical WPINN, with only $50,837$ parameters. Furthermore, PIQNN-(I, II) and classical PINN consistently fail to converge to meaningful solutions, exhibiting unstable training behavior and large errors.

Figure~\ref{KG-Exact-pred} provides a clear visual comparison of the exact and predicted solutions for the 1D Klein-Gordon equation using the proposed WPIQNN method. The results are shown for two cases with different frequencies, \( a = 5 \) and \( a = 10 \). In each case, the exact solution, WPIQNN prediction, and pointwise absolute errors are presented. It can be observed that WPIQNN accurately captures the oscillatory nature of the solutions in both scenarios, including the more challenging high-frequency case with \( a = 10 \). The error plots demonstrate that the errors remain consistently small throughout the domain, without noticeable deterioration near boundaries or regions of rapid oscillation. These visual results align well with the quantitative findings reported in Table~\ref{KG-table}, confirming the accuracy and robustness of WPIQNN for solving complex, highly oscillatory problems.


\begin{example}[\textbf{Maxwell equation}]\label{Example-4}
Maxwell’s equations form the cornerstone of classical electromagnetism, encapsulating the behavior of electric and magnetic fields and their interactions with materials. These equations are instrumental in understanding a wide array of physical phenomena and are fundamental in various technological domains. From wave propagation in communication systems to the principles of medical imaging and aerospace applications, Maxwell’s framework provides the mathematical backbone for electromagnetic theory.
In their differential form, Maxwell’s equations are expressed as:

\begin{equation}\label{General-maxwell}
\begin{cases}
\nabla \cdot \mathbf{E} = \dfrac{\rho}{\varepsilon}, \\[8pt]
\nabla \cdot \mathbf{B} =0,\\[8pt] 
\nabla \times \mathbf{E} = -\dfrac{\partial \mathbf{B}}{\partial t},\\[8pt]
\nabla \times \mathbf{H} = \mathbf{J} + \dfrac{\partial \mathbf{D}}{\partial t}
\end{cases}
\end{equation}
\end{example}
where, the electric displacement field $\mathbf{D} = \varepsilon \mathbf{E}$, and and the magnetic flux density $\mathbf{B} = \mu \mathbf{H}$. \(\mathbf{E}\) and  \(\mathbf{H}\) are electric and  magnetic fields respectively,  where \(\varepsilon_0\) is the electric permittivity and \(\mu_0\) is the magnetic permeability of the medium.

For a one-dimensional cavity model, we consider propagation in the $x-$direction assuming transverse electromagnetic waves with fields $E_y(x,t)$ and $H_z(x,t)$,  electric charge density $\rho = 0$, and electric current density $\mathbf{J} = 0$. Then the governing equation \eqref{General-maxwell} simplifies to
\begin{equation}\label{1d-maxwell}
\dfrac{\partial E_y}{\partial t} = -\dfrac{1}{\varepsilon(x)} \dfrac{\partial H_z}{\partial x}, \quad
\dfrac{\partial H_z}{\partial t} = -\dfrac{1}{\mu(x)} \dfrac{\partial E_y}{\partial x}, \qquad x \in [0,1],\ t \in [0,1].
\end{equation}
One of the major difficulties in solving Maxwell’s equations is the presence of rapid oscillations in space and time. These oscillations become even more challenging to handle when the wave propagates through heterogeneous media, where material properties such as permittivity ($\varepsilon$) and permeability ($\mu$) can change abruptly at interfaces. Such discontinuities introduce sharp variations in the fields, making it harder for standard PINNs to accurately approximate the solution. In this study, we address this challenge by solving Maxwell’s equations for both homogeneous and heterogeneous media. We compare the performance of traditional PINN and W-PINN with the proposed WPIQNN approach.\\[8pt]
\textbf{\large A. Homogeneous Media}\\[3pt]

In a homogeneous medium, \(\varepsilon(x) = \varepsilon\) and \(\mu(x) = \mu\) are constants. Applying perfectly electric conductor boundary conditions, we enforce:
\begin{equation}\label{homo-maxwell_bc}
    \begin{cases}
E_y(0,t) = E_y(1,t) = 0, \\[8pt]
\left.\dfrac{\partial H_z(x,t)}{\partial x}\right|_{x=0,1} = 0.\\[8pt]
\end{cases}
\end{equation}
An analytical solution in this scenario is:
\begin{equation}
    \begin{cases}
E_y(x,t) = \sin(n \pi x) \cos(\omega t), \\[8pt]
H_z(x,t) = -\cos(n \pi x) \sin(\omega t),
\end{cases}
\end{equation}
where \(n =4 \) is a mode number, and \(\omega = \dfrac{n \pi}{l}\) is the cavity frequency and $l$ is the length of medium.

For the homogeneous media case, the WPIQNN and WPINN architectures were implemented with the same network configuration to ensure consistency in comparison. Each model consisted of two hidden layers in the feature mapping network and four layers in the secondary neural network. To capture the behavior of Maxwell’s equations accurately, the number of collocation points was set to \(N_c = 2^{13}\), with \(N_0 = 500\) initial points and \(N_b = 500\) boundary points. Additionally, a wavelet resolution of \([-10,4]\) was employed in both cases.

\begin{table}[ht]
\centering
\caption{The performance of different methods Maxwell's equation ~\eqref{1d-maxwell} in homogenous media \eqref{homo-maxwell_bc}.}
\begin{tabular}{lccc}
\toprule
\multirow{2}{*}{\textbf{Method}} & \multicolumn{2}{c}{$\mathcal{L}_2$-error}  & \multirow{2}{*}{\textbf{Trainable parameters}} \\
\cmidrule(lr){2-3}
 & $E_y$  & $H_z$  \\
\midrule
Classical PINN & $3.08\pm 1.18\times 10^{-1}$ & $ 5.18\pm1.93\times 10^{-1}$ &$20,802$ \\
PIQNN-I \cite{xiao2024physics} & $5.87\pm 1.81\times 10^{-1}$ & $ 7.35\pm2.39\times 10^{-1}$ &$302 (7\times 6)$ \\
PIQNN-II \cite{xiao2024physics} & $4.12\pm 2.31\times 10^{-1}$ & $ 6.23\pm2.27\times 10^{-1}$ &$338(7\times 6)$ \\
W-PINN \cite{pandey2024efficient} & $ 6.34\pm2.12\times 10^{-4}$   &  $6.81\pm
 3.01 \times 10^{-4}$   &  $23,24,003$\\
WPIQNN &  $\mathbf{4.64\pm 2.15 \times 10^{-4}}$   & $\mathbf{4.36\pm 1.27 \times 10^{-4}}$  &$29,183$ \\
\bottomrule
\end{tabular}
\label{maxwell-homo-table}
\end{table}
The performance of the proposed WPIQNN framework for solving the one-dimensional Maxwell’s equation~\eqref{1d-maxwell} in homogeneous media~\eqref{homo-maxwell_bc} is comprehensively evaluated and compared against PIQNN-(I, II), classical PINN, and WPINN; the results are summarized in Table~\ref{maxwell-homo-table}. The results clearly demonstrate the superior performance of WPIQNN across all metrics. WPIQNN achieves the lowest relative \(\mathcal{L}_2\)-errors for both the electric field component (\(E_y\)) and the magnetic field component (\(H_z\)), outperforming the other methods by a notable margin. Beyond accuracy, WPIQNN offers significant efficiency in terms of trainable parameters compared to classical WPINN. While WPINN requires approximately $2.3$ million parameters to achieve its results, WPIQNN attains superior accuracy with just $29,183$ parameters, using approximately $2\%$ of the parameters required by WPINN. This drastic reduction highlights the efficiency of leveraging quantum-inspired architectures in physics-informed learning frameworks. Furthermore, PIQNN-(I, II) and classical PINN not only yield much larger errors but also exhibit unstable training behavior, particularly due to the oscillatory nature of solutions in electromagnetic problems. This reinforces the limitations of existing PIQNNs and conventional PINNs for such cases. In contrast, WPIQNN demonstrates stable convergence and consistently accurate solutions.

\begin{figure}[ht]
    \centering
    \includegraphics[width=0.85\linewidth]{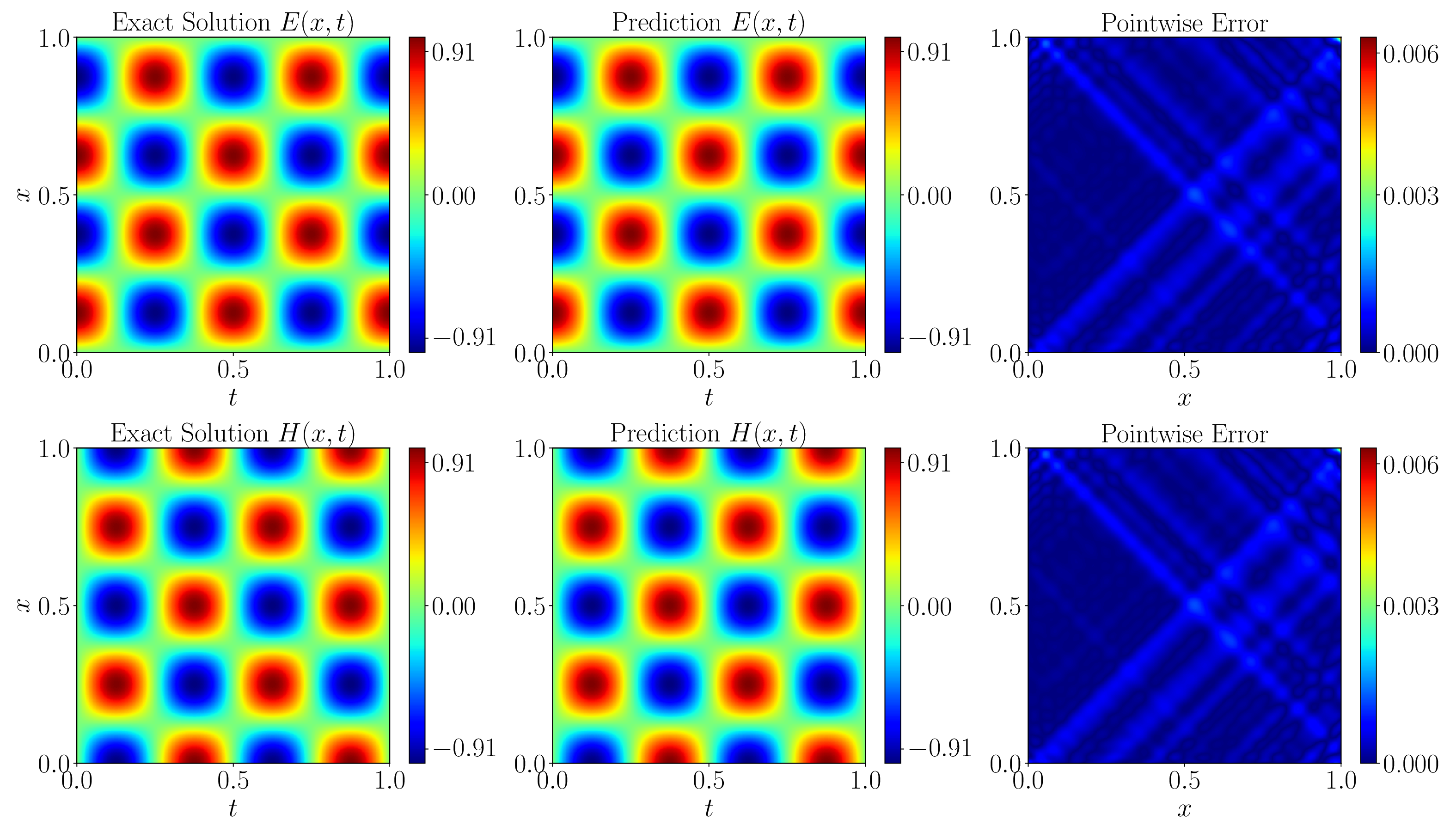}
    \caption{ Solutions to Maxwell's equation \eqref{1d-maxwell} with boundary condition \eqref{homo-maxwell_bc} in Homogeneous medium showing (left) exact solution, (center) WPIQNN prediction, and (right) pointwise absolute error.}
    \label{maxwell-homo-pw}
\end{figure}
Figure~\ref{maxwell-homo-pw} illustrates the comparison between the exact and predicted solutions for the one-dimensional Maxwell’s equation in a homogeneous medium, highlighting the performance of the proposed WPIQNN framework. The predictions for both the electric field \( E(x,t) \) and the magnetic field \( H(x,t) \) align closely with the exact solutions, successfully capturing the inherent oscillatory behavior of the electromagnetic fields. The pointwise absolute error remains consistently low across the entire spatio-temporal domain, indicating the model’s robustness and stability even in the presence of rapid oscillations. These results clearly demonstrate ability of the WPIQNN  to deliver highly accurate and stable solutions with significantly fewer trainable parameters compared to existing methods. The observed accuracy aligns well with the quantitative comparisons presented earlier, further confirming the advantage of WPIQNN in addressing challenging time-dependent wave propagation problems.\\[8pt]
\textbf{\large B. Heterogeneous media}\\[3pt]

For a heterogeneous medium, consider the domain split into two subdomains:
\(\Omega_1 = [0, 0.5]\) and \(\Omega_2 = [0.5, 1]\), with piecewise constant material properties:
\begin{equation}\label{hetero-maxwell_bc}
\varepsilon(x) = 
\begin{cases}
1, & x \in \Omega_1, \\
1.5, & x \in \Omega_2,
\end{cases}
\qquad
\mu(x) = 
\begin{cases}
1, & x \in \Omega_1, \\
4.5, & x \in \Omega_2.
\end{cases}
\end{equation}
Analytical solutions in this configuration are:
\begin{equation}
E_y(x,t) = 
\begin{cases}
\cos(2t - 2x + 1) + 0.5\cos(2t + 2x - 1), & x \in \Omega_1, \\
1.5\cos(2t - 3x + 1.5), & x \in \Omega_2,
\end{cases}
\end{equation}

\begin{equation}
H_z(x,t) = 
\begin{cases}
\cos(2t - 2x + 1) - 0.5\cos(2t + 2x - 1), & x \in \Omega_1, \\
0.5\cos(2t - 3x + 1.5), & x \in \Omega_2.
\end{cases}
\end{equation}
At interface \(x = 0.5\), the continuity of the tangential components of the fields yields the interface following conditions:
\begin{equation}
\left\{
\begin{aligned}
E_y(0.5^-,t) &= E_y(0.5^+,t), \\
H_z(0.5^-,t) &= H_z(0.5^+,t).
\end{aligned}
\right.
\end{equation}
For the heterogeneous media case, the WPIQNN and WPINN architectures used the same network configuration as in the homogeneous case, with two hidden layers in the feature mapping network and four layers in the secondary neural network. To accurately capture the discontinuities introduced by the piecewise constant material properties, the domain was divided into two subdomains, \(\Omega_1 = [0,0.5]\) and \(\Omega_2 = [0.5,1]\), and trained separately. The training employed \(N_c = 2^{13}\) collocation points in each subdomain, with \(N_0 = 500\) initial points and \(N_b = 500\) boundary points. Additionally, \(10{,}000\) interface points were introduced at \(x=0.5\) to enforce continuity conditions, and a refined wavelet resolution of \([-10,5]\) was used to resolve the local features more effectively.

\begin{table}[ht]
\centering
\caption{The performance of different methods Maxwell's equation~\eqref{1d-maxwell} in heterogenous media \eqref{hetero-maxwell_bc}.}
\begin{tabular}{lccc}
\toprule
\multirow{2}{*}{\textbf{Method}} & \multicolumn{2}{c}{$\mathcal{L}_2$-error}  & \multirow{2}{*}{\textbf{Trainable parameters}} \\
\cmidrule(lr){2-3}
 & $E_y$  & $H_z$  \\
\midrule
Classical PINN & $4.17\pm 2.19 \times 10^{-1}$   &  $3.17\pm 1.19 \times 10^{-1}$ & $41,604$ \\
PIQNN-I \cite{xiao2024physics} & $1.38\pm 2.55\times 10^{-3}$ & $2.75\pm 2.31\times 10^{-3}$ & $460 (5\times6)$ \\
PIQNN-II \cite{xiao2024physics} & $2.38\pm 1.36\times 10^{-3}$ & $3.54\pm 1.53\times 10^{-3}$ &$532 (5\times6)$ \\
WPINN \cite{pandey2024efficient} & $4.17\pm 2.19 \times 10^{-4}$   &  $3.17\pm 1.19 \times 10^{-4}$   & $46,48,006$ \\
WPIQNN &  $\mathbf{2.17\pm 0.95 \times 10^{-4}}$   & $\mathbf{1.70\pm 1.15 \times 10^{-4}}$  &  $58,366$ \\
\bottomrule
\end{tabular}
\label{tab:maxwell-hetero-sol}
\end{table}
Table~\ref{tab:maxwell-hetero-sol} compares the performance of the proposed WPIQNN framework with PIQNN-(I, II), Classical PINN, and classical WPINN for solving the one-dimensional Maxwell’s equation~\eqref{1d-maxwell} in heterogeneous media~\eqref{hetero-maxwell_bc}. Similarly to the homogeneous case, the results clearly demonstrate the superior performance of WPIQNN across all metrics. WPIQNN achieves the lowest relative \(\mathcal{L}_2\)-errors for both the electric field component (\(E_y\)) and the magnetic field component (\(H_z\)), outperforming other methods by a significant margin. In addition to improved accuracy, WPIQNN demonstrates remarkable efficiency in terms of trainable parameters. While WPINN requires more than $4.6$ million parameters to achieve its performance, WPIQNN attains even better results with only $58,366$ parameters, using only approximately $2\%$ of the parameters required by WPINN. The Classical PINN, on the other hand, suffers from large errors and unstable convergence due to the challenges posed by material discontinuities. These results further confirm that WPIQNN provides an effective balance between predictive accuracy and computational cost, particularly for challenging problems involving heterogeneous media.

\begin{figure}[ht]
    \centering
    \includegraphics[width=0.85\linewidth]{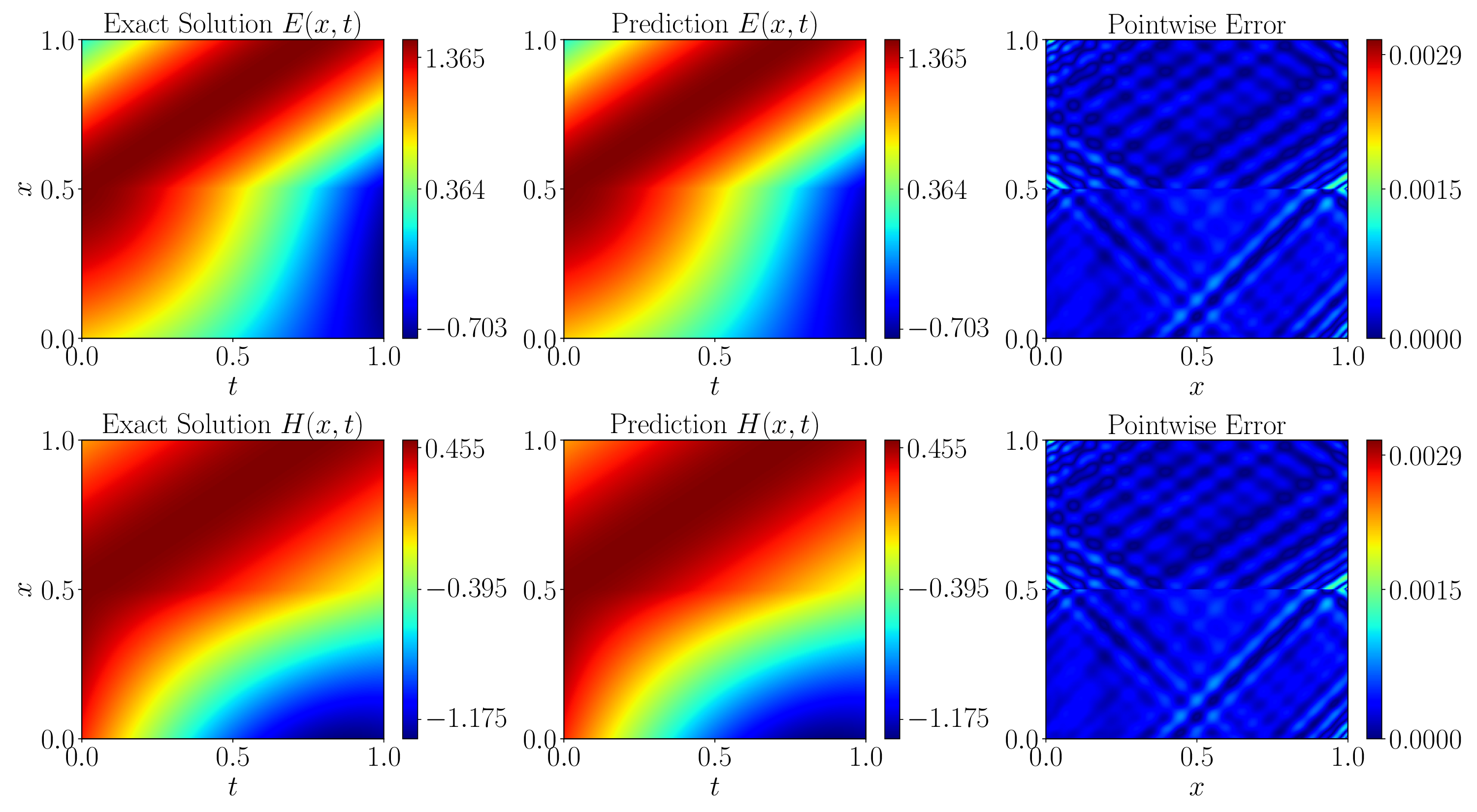}
    \caption{ Solutions to Maxwell's equation in heterogeneous medium showing (left) exact solution, (center) WPIQNN prediction, and (right) pointwise absolute error for the equation\eqref{hetero-maxwell_bc}.}
    \label{maxwell-hetero-pw}
\end{figure}
The performance of the proposed WPIQNN in solving Maxwell’s equation within a heterogeneous medium is illustrated through Figure~\ref{maxwell-hetero-pw}. The results clearly demonstrate that WPIQNN successfully captures the correct solution behavior across both subdomains, effectively handling the material discontinuities inherent in heterogeneous media. The predicted solutions for both electric and magnetic fields align closely with the exact solutions, thereby accurately preserving the oscillatory characteristics within each region. Moreover, the explicit treatment of interface points ensures that the continuity conditions at the material boundaries are properly enforced, as reflected in the pointwise absolute error plots. The errors remain uniformly low across the domain, with only slight localized deviations near the interfaces where sharp transitions occur. These findings further highlight the robustness and accuracy of WPIQNN in addressing electromagnetic wave propagation problems involving complex heterogeneous structures.

\section{Conclusion}\label{sec_concl}

We developed a WPIQNN framework that successfully addresses multiscale PDEs with sharp gradients, rapid oscillations, and stiff behavior, which are often challenging for traditional PINNs and PIQNNs. The proposed technique introduces a wavelet-based approach that combines the multiresolution capabilities of wavelet basis functions with the expressive power and parameter efficiency of QNNs. The wavelet transform ensures a balanced loss representation across different scales, whereas the QNN component substantially reduces the number of trainable parameters and offers an improved effective dimension. Moreover, by leveraging analytically computable wavelet derivatives, the WPIQNN eliminates the need for automatic differentiation, thereby avoiding the computational overhead of backpropagation and achieving a three to five times speedup in training time compared to existing PIQNNs. Extensive numerical experiments demonstrate that WPIQNN achieves better accuracy using less than five percent of the parameters required by classical W-PINNs, highlighting its efficiency without compromising accuracy. In short, the WPIQNN offers an efficient, reliable, and flexible way to solve complex problems involving multiple scales. The proposed method overcomes the key weaknesses of current models and opens new possibilities for faster and more accurate scientific computing using quantum technology.

Future work can focus on solving more complex PDEs from fluid dynamics and electromagnetics, involving multiscale features, complex geometries, and irregular sampling, where the WPIQNN framework has the potential to offer significant advantages.

\newpage
\begin{appendix}\label{appendix}
\section{Appendix}
\subsection{Parameters used for different examples}
\begin{table}[ht!]
\centering
\caption{Parameters used for solving different PDE examples using the WPIQNN framework}
\renewcommand{\arraystretch}{1.2}
\resizebox{\textwidth}{!}{%
\begin{tabular}{|c|ccc|c|cc|cc|}
\hline
\multirow{2}{*}{\textbf{Parameters}} & \multicolumn{3}{c|}{\textbf{Heat Conduction}\ref{Example-1}} & \multirow{2}{*}{\textbf{Helmholtz}\ref{Example-2}} & \multicolumn{2}{c|}{\textbf{Klein-Gordon}\ref{Example-3}} &  \multicolumn{2}{c|}{\textbf{Maxwell's Equation}\ref{Example-4}} \\[6pt] 
 & $\varepsilon=0.5$ & $\varepsilon=0.25$ & $\varepsilon=0.15$ & & $a=5$ & $a=10$ & Heterogeneous & Homogeneous \\[4pt] \hline
\makecell{Domain}      &\multicolumn{3}{c|}{$[-1,1]\times[0,1]$}  & $[-1,1]\times[-1,1]$ & \multicolumn{2}{c|}{$[0,1]\times[0,1]$} & \multicolumn{2}{c|}{$[0,1]\times[0,1]$}   \\[6pt]  \hline
\makecell{No. of Collocation \\ Points $(N_c)$} & $2^{13}$ & $2^{13}$ & $2^{13}$ & $2^{14}$ & $2^{13}$ & $2^{14}$ & $2^{13}$ & $2^{13}$   \\[8pt]  \hline
\makecell{No. of Boundary \\ Points $(N_b)$}    & $10^3$   & $10^3$   & $2\times10^3$ & $10^3$ & $5\times10^2$ & $5\times10^2$ & $5\times10^2$ &$5\times10^2$   \\[8pt]  \hline
\makecell{No. of Initial \\ Points $(N_0)$}     & $10^3$   & $10^3$   & $10^3$ & --- &  $500$ & $500$ & $500$ & $500$   \\[8pt]  \hline

\makecell{No. of training \\ Epochs}     & $5\times10^4$   & $5\times10^4$   & $5\times10^4$ & $5\times10^4$ &  $1.2\times10^5$ & $4\times10^5$ & $5\times 10^{4}$ & $5\times 10^{4}$   \\[8pt]  \hline

\makecell{Initial Learning  \\Rate}     & $10^{-2}$   & $10^{-2}$   & $10^{-2}$ & $10^{-2}$ &  $10^{-2}$ & $10^{-2}$ & $10^{-2}$ & $10^{-2}$   \\[8pt]  \hline
\makecell{Scheduler}     & Yes   & Yes   & Yes & Yes &  Yes & Yes & Yes & Yes   \\[8pt]  \hline
\makecell{Set of \\ Resolution}                    & $[-5,4]$ & $[-5,4]$ & $[-5,5]$ & $[-4,5]$ &  $[-10,4]$ & $[-10,5]$ & $[-10,4]$ & $[-10,5]$   \\[8pt]  \hline
\makecell{No. of Qubits \\ (QNN1)}              & $4$      & $4$      & $4$ & $4$  & $4$  & $4$  &  $4$  & $4$    \\[8pt]  \hline
\makecell{No. of Layers \\ (QNN1)}              & $2$      & $2$      & $2$ & $2$  & $2$  & $2$  &  $2$  & $2$    \\[8pt]  \hline
\makecell{No. of Qubits \\ (QNN2)}              & $13$     & $13$     & $13$& $14$ & $13$ & $14$ &  $13$ & $13$   \\[8pt]  \hline
\makecell{No. of Layers \\ (QNN2)}              & $4$      & $4$      & $4$ & $4$  & $4$  & $4$  &  $4$  & $4$    \\[8pt]  \hline
\end{tabular}%
}
\label{parameters-table}
\end{table}
\end{appendix}
\section*{Declarations}
\subsection*{Ethical Approval}
The authors confirm that the present work adheres to the ethical guidelines of the journal. This work has not been published elsewhere and is not under consideration by any other publication.
\subsection*{Funding}
The authors did not receive any specific funding for this work.

\subsection*{Conflict of Interest}
The authors declare that they have no conflicts of interest.
\subsection*{Author Information}
Deepak Gupta${}^a$, Himanshu Pandey${}^a$, Ratikanta Behera${}^a$ \\[0.5em]
${}^a$Department of Computational and Data Sciences, Indian Institute of Science, Bangalore, India \\[0.5em]
\textbf{Corresponding author:} Ratikanta Behera

		
\section*{Acknowledgements}
The authors would like to acknowledge the Department of Computational and Data Sciences, Indian Institute of Science, Bangalore, for providing the research environment and facilities necessary to carry out this work. 

\section*{Author Contributions}
\begin{itemize}
    \item \textbf{Deepak Gupta:} Writing - original draft, investigation, methodology, coding. 
    \item \textbf{Himanshu Pandey:} Reviewing-manuscript and investigation, coding, visualization with validation.
    \item \textbf{Ratikanta Behera:} Supervision, visualization, validation, writing review and editing.
\end{itemize}

\bibliographystyle{abbrv}
\bibliography{WPIQNN-Ref}	
\end{document}